\documentclass[11pt]{article}

\usepackage[preprint]{acl}

\usepackage{times}
\usepackage{latexsym}
\usepackage{tabularx,booktabs}
\usepackage{array}
\usepackage{adjustbox}
\usepackage{xcolor}
\usepackage[table]{xcolor}
\usepackage{amsmath}
\usepackage{amssymb}
\usepackage{tikz}
\usepackage{longtable}
\usepackage{multirow}
\usetikzlibrary{arrows.meta,positioning,matrix,fit,calc}

\usepackage[T1]{fontenc}

\usepackage[utf8]{inputenc}

\usepackage{microtype}

\usepackage{inconsolata}

\usepackage{graphicx}
\usepackage[most]{tcolorbox}
\usepackage{fontawesome5} 
\usepackage{enumitem}
\setlist[itemize]{noitemsep, topsep=0pt, leftmargin=*}

\newlist{compactitem}{itemize}{1}
\setlist[compactitem]{noitemsep, topsep=0pt}

\newtcolorbox{promptbox}[2][]{
  enhanced,
  colback=white!98!blue!2, 
  colframe=blue!70!black,  
  coltitle=white,          
  fonttitle=\bfseries\sffamily,
  title={\faTasks[regular]\hspace{1mm}~#2},
  left=2mm, right=2mm, top=1mm, bottom=1mm,
  arc=3mm,                  
  attach boxed title to top left={xshift=3mm, yshift*=-2mm},
  boxed title style={
    colback=blue!80!black,
    size=small,
    sharp corners=south,
    bottom=0.5mm, top=0.5mm,
    left=1mm, right=1mm,
    fontupper=\bfseries\sffamily,
    boxrule=0pt,
    drop shadow
  },
  boxrule=0.9pt,
  drop shadow southeast,
}

\title{GRASP: \underline{GR}anularity-\underline{A}ware \underline{S}earch \underline{P}olicy for Agentic RAG}

\author{
 \textbf{Varun Gandhi\textsuperscript{1,*}},
 \textbf{Jaewook Lee\textsuperscript{1,*}}
 \textbf{Shantanu Todmal\textsuperscript{1}},
  \\
 \textbf{Franck Dernoncourt\textsuperscript{2}},
 \textbf{Ryan Rossi\textsuperscript{2}},
 \textbf{Zichao Wang\textsuperscript{2}},
 \textbf{Andrew Lan\textsuperscript{1}},
\\
 \textsuperscript{1}University of Massachusetts Amherst,
 \textsuperscript{2}Adobe Research
\\
 \textsuperscript{*}Equal contribution.
\\
 \small{
   \textbf{Correspondence:} \href{mailto:vgandhi@umass.edu}{vgandhi@umass.edu}
 }
}

\begin{document}
\maketitle
\begin{abstract}
Agentic retrieval-augmented generation (RAG) extends static RAG by allowing language models to iteratively reason, generate search queries, retrieve evidence, and predict answers. However, it remains challenging for models to decide when to retrieve, whether to use lexical matching or semantic similarity, and how to control context granularity to prevent irrelevant tokens from interfering with agent reasoning. In this paper, we introduce GRASP, a reinforcement learning (RL) framework for training agents to adaptively coordinate complementary retrieval tools during multi-step reasoning. GRASP provides the agent with semantic search, keyword search, and paragraph-reading actions, enabling it to retrieve sentence-level evidence and expand further context only when needed. We train the policy with a reward that jointly accounts for answer accuracy, grounded reading, complementary search, and turn efficiency. Experiments on multi-hop reasoning benchmarks show that GRASP improves both retrieval recall and downstream question answering performance compared with single-step retrieval, prompting-based agentic RAG, and RL-based retrieval baselines. Qualitative and ablation analyses show that the learned policy develops interpretable skimming and scanning behavior: it uses semantic search for broad exploration, paragraph reading for local verification, and keyword search for entity-specific evidence. These results suggest that learning to coordinate retrieval signals and context granularity is critical for agent's correct reasoning.
\end{abstract}

\section{Introduction}
\label{sec:intro}
Large language models (LLMs) have demonstrated strong capabilities in understanding and generating natural language, enabling document-centered tasks such as long document summarization, information extraction, and reasoning for question answering~\cite{perot2024lmdx,jiao2023instruct,li2024contradoc}. However, their parametric knowledge is largely limited to information acquired during pretraining, which can lead to factual inaccuracies and limited adaptability when up-to-date or domain-specific knowledge is required~\cite{li2025knowledge,wang2024factuality,nie2025embedding}. Retrieval-Augmented Generation (RAG) has emerged as a mainstream approach to addressing this limitation by incorporating external knowledge at inference time.

Traditional, \emph{static} RAG is typically formulated as a static, single-step pipeline: A retriever first identifies relevant chunks from external documents, then an LLM generates a response conditioned on the retrieved evidence. Here, a chunk refers to a smaller textual unit such as a passage, paragraph, or fixed-length span, obtained by segmenting a long document, serving as the basic unit of retrieval. The retriever ranks chunks according to their relevance to the input query, using either \textit{lexical matching} methods, such as BM25, or \textit{semantic similarity} methods, such as dense retrieval models that encode queries and passages into continuous vector representations~\cite{karpukhin2020dense}. In practice, the initially retrieved chunks are often further refined by a re-ranker, such as a cross-encoder, which computes relevance scores between query and chunk to improve final retrieval results~\cite{ren2021rocketqav2}.

\begin{figure}[t!]
    \centering
    \includegraphics[width=1\linewidth]{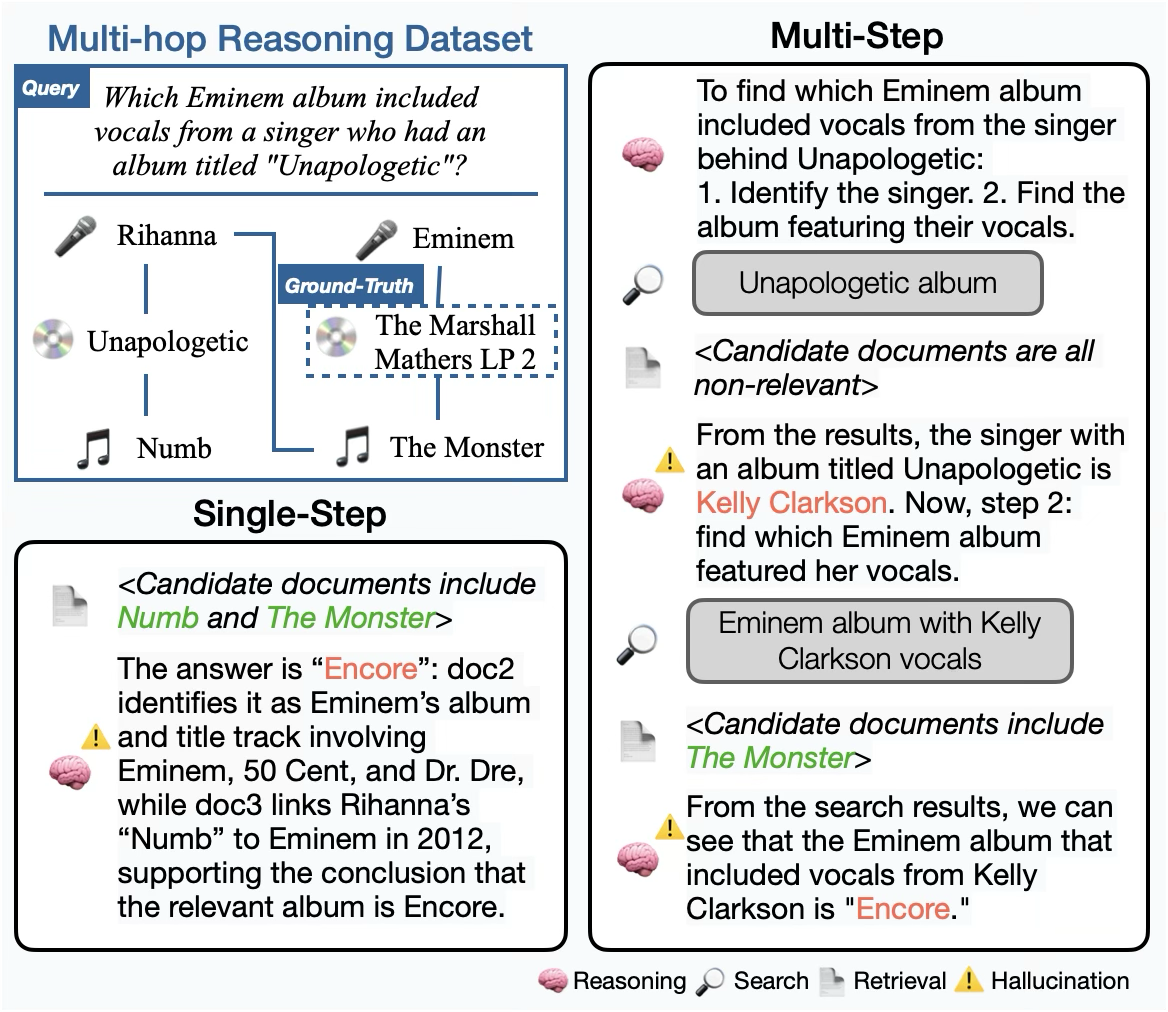}
    \caption{Example of why carefully controlled context is critical for multi-hop reasoning. Single-step reasoning fails despite sufficient evidence because irrelevant content distracts the model, while multi-step reasoning compounds errors by retrieving non-relevant documents and hallucinating a misleading connection.}
    \label{fig:motivation}
\end{figure}

Recently, RAG has evolved from static pipelines into \emph{agentic} RAG, where the LLM acts as an agent that iteratively reasons, generates queries, retrieves evidence, and synthesizes answers over multiple steps. This evolution changes the what the retriever does from simply ranking a fixed set of candidate chunks to deciding how retrieval should happen at each step. In particular, agentic RAG introduces three challenges: (1) leveraging complementary retrieval signals of lexical matching and semantic similarity, (2) selecting the appropriate context granularity to avoid filling the context window with irrelevant tokens, and (3) how to train policies that coordinate multiple tools, and decide when, where, and how to retrieve across multiple reasoning steps.

Figure~\ref{fig:motivation} illustrates why these challenges are critical for multi-hop reasoning in both static and agentic RAG. In static RAG, complementary retrieval signals can be incorporated relatively easily using a hybrid retriever. However, even when the correct documents are retrieved, the model may still hallucinate if the retrieved context is too coarse-grained, noisy, or distractor-heavy (e.g., the retriever retrieves the two documents needed to answer the question, but the reasoning process hallucinates and generates an incorrect answer). Large chunks can obscure the specific evidence needed for reasoning, causing the model to attend to irrelevant information or fail to complete the reasoning chain. Moreover, because static RAG typically performs retrieval only once, it cannot adapt its retrieval behavior based on intermediate reasoning states.

Agentic RAG partially addresses this limitation by allowing the model to retrieve evidence iteratively. However, it inherits the same hallucination issue when retrieved chunks are coarse-grained or noisy, as distracting context can still dilute the evidence needed for each reasoning step (e.g., the agent hallucinates Kelly Clarkson from irrelevant documents, even though the name is not mentioned in the retrieved documents). This problem is amplified in agentic settings because reasoning proceeds incrementally: an incorrect retrieval decision or flawed intermediate conclusion can propagate through later steps and derail the trajectory (e.g., the final reasoning produces an incorrect answer and also identifies the wrong singer). 
Furthermore, leveraging complementary retrieval signals is no longer as simple as applying a fixed hybrid retriever, as retrieval needs are spread across the reasoning trajectory: some steps require lexical matching to retrieve entity-specific evidence, while others require semantic search to connect conceptually related information. Therefore, learning effective retriever-tool policies that decide when to retrieve, which retrieval signal to use, and what granularity of evidence to return at each reasoning step becomes a key challenge in agentic RAG. Existing approaches typically use a single retriever or a fixed, coarse-grained chunk granularity when training such policies, limiting their ability to adapt retrieval decisions across different stages of multi-step reasoning~\cite{jin2025search,luo2025marag}.

In this paper, we study how to train agentic RAG systems to use retrieval tools effectively and limit irrelevant retrieved docs in the context window. Unlike standard RAG settings that retrieve a fixed set of passages upfront, our setting requires agents to decide when to search, which retrieval strategy to use, and how to manage retrieved information in context window while accumulating evidence. 
To address this challenge, we introduce \textbf{GRASP}, an RL framework that trains language models to dynamically coordinate complementary retrieval tools and adaptively retrieve further context of relevant documents. We use \texttt{Qwen/Qwen2.5-3B-Instruct}~\cite{qwen2.5} as the target policy model, providing a setting in which retrieval granularity and evidence selection play a central role in successful multi-hop reasoning.
%

\textbf{Contributions.} First, we formulate adaptive retrieval tool selection for agentic RAG as an RL problem, enabling the model to learn when to invoke lexical search, semantic search, and context expansion during multi-step reasoning. Second, we introduce a reward design that jointly accounts for  answer accuracy, grounded reading, complementary search, and turn efficiency, guiding the agent to retrieve sufficient evidence while controlling both tool usage and context granularity. Third, we evaluate the learned policy against prompting-based and RL-based baselines and show that it achieves higher answer accuracy. Finally, we analyze the policy’s emergent behavior and show that it resembles human information foraging.

\section{Problem Formulation}

In this section, we define an agentic RAG framework where the agent can dynamically retrieve, manage, and compose evidence from multiple retrieval signals and context granularities. Let the query be $q$, and let the retrieval corpus be $\mathcal{D}=\mathcal{D}_{g}\cup\mathcal{D}_{d}$, where $\mathcal{D}_{g}$ and $\mathcal{D}_{d}$ denote gold, i.e., paragraphs containing ground-truth supporting evidences to answer $q$ and distractor paragraphs, respectively. Each paragraph $d\in\mathcal{D}$ is decomposed into sentences $d=(s_1,\dots,s_n)$, forming a sentence-level corpus $\mathcal{S}=\bigcup_{d\in\mathcal{D}} d$. Each sentence $s\in\mathcal{S}$ is associated with representations that expose complementary retrieval signals, including lexical and semantic information, while a mapping $f:\mathcal{S}\rightarrow\mathcal{D}$ indicates which paragraph each sentence comes from. Given $q$, the objective of the framework is to generate an answer $\hat{y}$ by selecting and composing sufficient evidence from \(\mathcal{D}_{g}\) while avoiding reliance on distractor evidence from \(\mathcal{D}_{d}\).
%



\section{Methodology}
\label{sec:method}

\begin{figure*}
\centering
\includegraphics[width=1\linewidth]{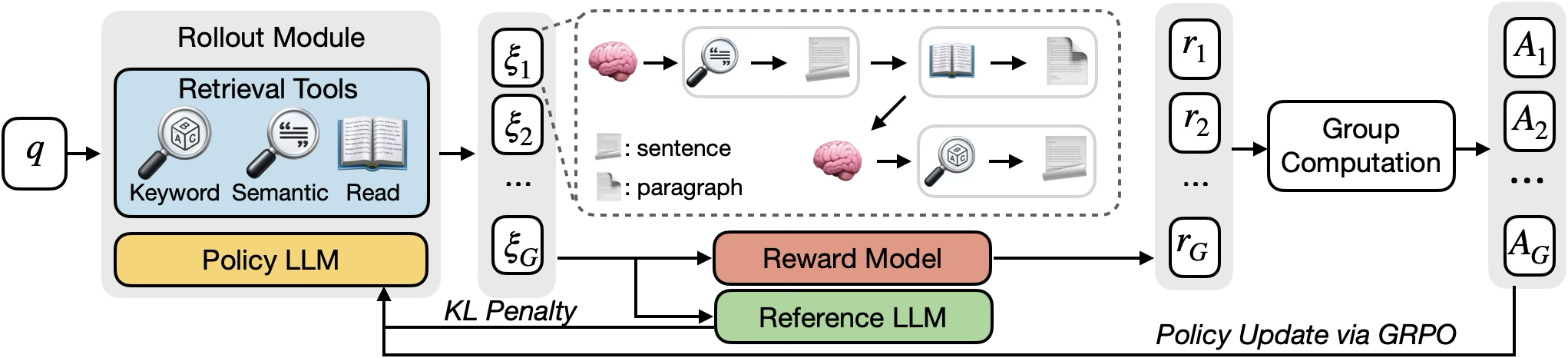}
\caption{Overview of the GRASP framework. Given a query $q$, the policy LLM generates multiple rollouts; for example, trajectory $\xi_1$ follows a reason $\rightarrow$ semantic search $\rightarrow$ read parent paragraph $\rightarrow$ reason $\rightarrow$ keyword search sequence. Each rollout is scored by a reward model, with a reference LLM providing a KL-regularization penalty. The policy is updated via GRPO using advantages computed from group-wise rewards.
}
\label{fig:method}
\end{figure*}

We formulate agentic RAG as a finite-horizon Markov decision process:
\[
\mathcal{M}=(\mathcal{X},\mathcal{A},P,R,T),
\]
where \(\mathcal{X}\) is the state space, \(\mathcal{A}\) is the action space, \(P\) is the transition function induced by agent generations, tool execution, and context updates, \(R\) is the trajectory-level reward, and \(T\) is the maximum number of turns in this iterative process.

At step \(t\), the state \(x_t\in\mathcal{X}\) is characterized by information available to the agent before it makes the next decision.
In our setting, \(x_t\) consists of the query \(q\), the interaction history \(h_{t-1}\), previously retrieved evidence, previously invoked tools, and the remaining turn budget, i.e., \(T-t+1\).
The interaction history contains both policy-generated tokens and environment-generated observations.

The policy LLM \(\pi_\theta\) (or agent) first generates a reasoning segment \(g_t\).  It then invokes a tool action $a_t$, and the retrieval environment returns the corresponding observation $o_t$. We update the interaction history $h_t$ by concatenating these components:
\[
g_t \sim \pi_\theta(\cdot \mid x_t), \quad
a_t \sim \pi_\theta(\cdot \mid x_t,g_t), \quad
a_t\in\mathcal{A}.
\]
\[
h_t = h_{t-1}\oplus g_t\oplus a_t\oplus o_t,
\]
The next generation then conditions on the updated history \(h_t\). 
Next, we present our action design, the reward design, the learning algorithm, and the implementation details.

\subsection{Action Design} \label{sec:method-action-design}

We define the action space around three evidence-acquisition tools:
\textbf{semantic search} $(\tau_s)$, \textbf{keyword search} $(\tau_k)$, and \textbf{read paragraph} $(\tau_r)$.
The first two retrieve sentence-level evidence from the corpus $\mathcal{S}$, while the third lets the agent retrieve the paragraph from which these sentences are drawn.
%
%
At each retrieval step $t$, after generating the reasoning segment $g_t$, the agent selects an action:
\[
a_t \in \mathcal{A}=\{\tau_s,\tau_k,\tau_r,\tau_{\mathrm{a}}\},
\]
where $\tau_{\mathrm{a}}$ denotes termination with a final answer.

Given a query $q_t$ generated by the agent, both retrieval actions return the top-$K$ sentences together with their parent paragraph identifiers:
\[
o_t \!=\! \tau(q_t)
      \!=\! \{(s_i, f(s_i), \mathrm{score}_i)\}_{i=1}^{K},
      \, \!\tau \!\in \!\{\tau_s,\tau_k\}.
\]
where $s_i\in\mathcal{S}$ is a retrieved sentence and $\mathrm{score}_i$ is the retrieval score induced by the selected retriever.
The two retrieval actions differ only in how this score is computed:
$\tau_s$ uses dense representations to capture semantic similarity, whereas $\tau_k$ uses lexical matching to favor exact or near-exact overlap.
Thus, semantic search can recover relevant evidence under lexical mismatch, while keyword search provides complementary lexical signals.

After observing sentence-level evidence in $o_t$, the agent proceeds with reasoning and may additionally invoke retrieval again or use the read action $\tau_r$ to access broader context, where for a retrieved sentence $s\in\mathcal{S}$, $\tau_r(s)=f(s)=d$ returns its parent paragraph $d\in\mathcal{D}$.

Finally, the agent uses the action $\tau_{\mathrm{a}}$ to produce the final answer $\hat{y}$, which is the agent's predicted answer to the original query $q$.


\subsection{Reward Design}
\label{sec:reward}
We formulate the trajectory-level reward to optimize both the quality of the final answer and the agent's intermediate evidence-seeking behavior, including grounded reading, complementary search, and turn efficiency.

\paragraph{Answer Accuracy ($R_A$)} This reward evaluates answer correctness by comparing the model's predicted answer to the reference answer. We calculate token-level F1 between predicted answer $\hat{y}$ and ground-truth answer $y$: $R_A=\mathrm{F1}(\hat{y},y)\in[0,1]$.

\paragraph{Grounded Reading ($R_R$)}
This reward evaluates whether the agent invokes the read paragraph tool $\tau_r$ on retrieved sentences from gold paragraphs rather than distractors. Let $\mathcal{D}_r$ be the set of documents associated with sentences expanded by $\tau_r$, and let $\mathcal{D}_g$ be the gold document set. We compute:
\[
R_R=\mathrm{F1}(\mathcal{D}_r,\,\mathcal{D}_g)\in[0,1].
\]
In other words, we reward reads of gold-document evidence and penalize unnecessary reads.

\paragraph{Complementary Search ($R_S$)} This reward evaluates whether both search tools retrieve at least one gold document from $\mathcal{D}_g$. Let $\mathcal{D}_s$ and $\mathcal{D}_k$ be the documents retrieved by $\tau_s$ and $\tau_k$:
\[
R_S\;=\;\mathbf{1}\!\left[\,\mathcal{D}_s\cap\mathcal{D}_g\neq\varnothing \;\wedge\; \mathcal{D}_k\cap\mathcal{D}_g\neq\varnothing\,\right].
\]
We set the reward only when both semantic and lexical search retrieves gold evidence. Doing so encourages the use of complementary retrieval signals while avoiding cases where the model relies on repeated calls to a single search tool.

\paragraph{Turn Efficiency ($R_E$)} This reward evaluates the agent's efficiency on the query by calculating how many turns it used. We define:
\[
R_E \;=\; \frac{T-T_{\text{cur}}}{T} \cdot \mathbf{1}[R_A > 0.5],
\]
where $T_{\text{cur}}$ is the number of turns used. This reward is given only when the answer reward exceeds $0.5$, preventing the agent from receiving efficiency credit through early guessing.

\paragraph{Total Reward ($R$)} We aggregate the final trajectory-level rewards as:
\[
R = R_A + \alpha \, R_R + \beta \, R_S + \gamma \, R_E.
\]
The total reward to have a maximum value of \(2.0\): \(1.0\) from the answer reward and \(1.0\) from the auxiliary rewards. We allocate the auxiliary reward weight as \(\alpha=0.7\), \(\beta=0.15\), and \(\gamma=0.15\). 




We allocate most of the auxiliary weight to $R_R$ to encourage reading and using relevant evidence under a limited context window, making it the primary signal for correct answering. We assign smaller weights to $R_S$ and $R_E$ to balance search quality and efficiency; higher search rewards could cause unnecessary retrieval, while higher efficiency rewards might discourage gathering enough evidence. These lower weights encourage effective searching and concise trajectories without distracting from the main goal. 
Importantly, by keeping $R_A$ unweighted and limiting the auxiliary terms to $1.0$ in aggregate, we ensure that answer accuracy remains the main driver of the learned policy. Because the auxiliary rewards guide \emph{how} the agent reasons and retrieves evidence, this design helps guard against reward hacking, where a policy learns to maximize easier intermediate signals while degrading final task performance in predicting the correct answer.
%




\subsection{Learning Algorithm}
\label{sec:method-learning-alg}

We learn the retrieval policy using Group-Relative Policy Optimization (GRPO)~\cite{shao2024deepseekmath}. Given a query \(q\), the policy induces a trajectory distribution \(p_\theta(\xi \mid q)\) through its interaction with the retrieval environment, from which we sample a group of \(G\) multi-turn trajectories:
\[
\xi_i \sim p_\theta(\xi \mid q), \qquad i=1,\ldots,G.
\]
Each trajectory $\xi$ records the agent’s states, retrieval decisions, and environment observations over \(T\) turns. We assign a trajectory-level reward \(R_i = R(\xi_i)\) based on the predicted answer, the documents retrieved or read by the agent, and the number of turns used. GRPO then updates the policy by comparing the \(G\) sampled trajectories relative to one another for the same query, rather than relying on an absolute reward scale.

\subsection{Implementation Details}
\label{sec:implementation}

Each training example from the HotpotQA distractor split consists of a question, an answer, sentence-level supporting-fact annotations, and two gold paragraphs containing the supporting facts together with eight distractor paragraphs. We build a global retrieval corpus where each training example has two gold documents and eight distractors. The corpus is indexed at the sentence level, with parent paragraph and document-title metadata retained for paragraph expansion and evidence tracking.

The system prompt instructs the model to interleave reasoning segments with calls to the three retrieval tools $\tau_s$, $\tau_k$, and $\tau_r$.
The model produces reasoning segment in \texttt{\textless{}think\textgreater{}} blocks, invokes the selected retrieval action, and returns the final answer using \texttt{\textless{}answer\textgreater{}} tags; the complete prompt is given in Appendix Table~\ref{tab:our_prompt}. Full data statistics, optimization hyperparameters, and infrastructure details are reported in Appendix~\ref{app:training} (Table~\ref{tab:train-config}).


We mask out the tool observation, i.e., tokens of retrieved documents, from the policy loss following \citet{jin2025search}, making sure that the model is trained only on tokens it generates itself, such as reasoning steps, tool calls, and final answers, rather than on text returned by the retriever.

\paragraph{Reward training progression.} The reward trajectory reported in Appendix Figure~\ref{fig:reward-progression} shows that the per-step total reward $R$ increases from approximately $0.26$ at the start of training to approximately $1.22$ at convergence. This trend indicates steady improvement under the aggregated answer-accuracy, grounded-reading, complementary-search, and turn-efficiency objectives defined in Section~\ref{sec:reward}.

\section{Experimental Setup}
\subsection{Dataset}
We use HotpotQA~\cite{yang2018hotpotqa}, 2WikiMultiHopQA~\cite{ho2020constructing}, and MuSiQue~\cite{trivedi2022musique} as multi-hop question answering benchmarks. Because the original test splits of these datasets are not publicly released with gold annotations, we follow standard practice and use each dataset's validation split as our held-out test set. Our main model is trained on HotpotQA and evaluated on all three datasets, enabling us to measure both in-domain performance and cross-dataset generalization. For each dataset we sample $500$ questions uniformly at random from the validation split (seed~$42$) and the agent retrieves from the full validation-split retrieval corpus, which we build by extracting the unique supporting and distractor paragraphs across every validation question. Appendix~\ref{app:datasets} lists the Hugging Face dataset identifiers and the corresponding retrieval-corpus sizes for all three benchmarks.

\subsection{Retriever}
We use \texttt{BM25}~\cite{robertson2009probabilistic} as the lexical retriever and \texttt{Qwen3-0.6B}~\cite{qwen3technicalreport} as the semantic retriever. We set the retrieval depth to top-$k=5$ for all methods.

\subsection{Baselines}
We compare our method to the following baselines to examine whether challenges introduced in Section~\ref{sec:intro}, namely (1) leveraging complementary retrieval signals, (2) adapting context granularity, and (3) learning adaptive retrieval decisions are addressed. See Appendix~\ref{sec:related_work} for a more detailed review of prior work.

\paragraph{Single-Step Retrieval}
We first consider single-step retrieval baselines following \citet{jin2025search}. While the original setting uses a semantic retriever, we extend the retriever space to lexical, semantic, and hybrid retrieval. For hybrid retrieval, we retrieve 50 candidates from each retriever, merge and deduplicate them, rerank the candidate pool with \texttt{Qwen3-Reranker-0.6B}~\cite{qwen3embedding}, and keep the final top-$k=5$ results. The hybrid baseline tests whether complementary lexical and semantic signals improve retrieval within a single step, whereas our method examines whether such complementary signals can be exploited more effectively across multiple retrieval steps.

\paragraph{Multi-Step Retrieval}
We next consider agentic RAG baselines, including both prompting-based and RL-based methods. For prompting-based agentic RAG, we compare against \textbf{IRCoT}~\cite{trivedi2023interleaving}, which interleaves retrieval with Chain-of-Thought (CoT) reasoning to guide subsequent retrieval steps. We use \texttt{gpt5-mini}~\cite{openai2026gpt5mini} to generate CoT traces, enabling us to evaluate whether our agent, without distillation from such a proprietary model, can achieve comparable performance. For RL-based agentic RAG, we compare against \textbf{Search-R1}~\cite{jin2025search}, using the existing checkpoint on Hugging Face~\footnote{https://huggingface.co/collections/PeterJinGo/search-r1}.

For the retriever in each agentic RAG baseline, we follow the original settings: IRCoT uses a lexical retriever, whereas Search-R1 uses a semantic retriever, and both methods retrieve evidence at the paragraph level. We also report a \textbf{Base} setting, which uses GRASP’s prompt and three tools for question answering using the base Qwen 2.5 3B Instruct model. This setting assesses the impact of RL training in our method.

\subsection{Evaluation Metrics}

\paragraph{Retrieval}
Before evaluating generation quality across methods, we first evaluate the quality of the retrieved documents to verify whether the agent is provided with sufficient evidence for downstream reasoning and answer generation. The single-step results show whether the complementary search signal is indeed helpful and indicate the lower bound of multi-step agentic RAG. In the multi-step setting, the metrics help evaluate whether the intermediate queries generated by the agent are valid. Since these tasks generally involve two-hop reasoning, a recall score of $0.5$ indicates that only one relevant passage was retrieved, which is insufficient. To calculate recall in the multi-step setting, we aggregate all passages retrieved from the queries and compute recall over the combined set.

\paragraph{QA} We evaluate generation quality using three metrics: (1) Exact Match (EM), (2) token-level F1 (F1), and (3) LLM-as-a-judge (JD). We use EM and F1 to measure lexical overlap between the generated answer and the gold answer, providing exact and approximate token-level assessments of answer correctness, respectively. To capture more semantically flexible notions of correctness over lexical overlap, we additionally use an LLM-as-a-judge metric with \texttt{gpt5-mini}. We use the judge to evaluate whether the generated answer contains the key information from the gold answer, is factually consistent with it, and avoids contradictions. The output is binary, indicating whether the generated answer is correct or incorrect. (See Appendix Table~\ref{tab:judge-prompt} for the prompt.)


\section{Results}
In this section, we first report quantitative retrieval and QA performance, then analyze the learned retrieval behavior qualitatively, and finally conduct ablation studies to evaluate the contribution of complementary search tools and context granularity.

\subsection{Quantitative Results}

\begin{table}[t!]
    \centering
    \small
\scalebox{.89}{
    \begin{tabular}{c c c c c}
        \toprule
        Method & Retriever & HotpotQA & 2Wiki & MuSiQue \\
        \midrule

        \multirow{3}{*}{Single-step} 
& Lexical & 0.61 & 0.49 & 0.35 \\
& Semantic  & 0.73 & 0.61 & 0.47 \\
& Hybrid & 0.86 & 0.60 & 0.59 \\
        \midrule

        IRCoT & Lexical & \textbf{0.91} & \underline{0.83} & \underline{0.62} \\
        Search-R1 (P)& \multirow{2}{*}{Semantic} & 0.76 & 0.74 & 0.55 \\
        Search-R1 (G)&  & 0.74 & 0.77 & \underline{0.62} \\
        Base & \multirow{2}{*}{Hybrid} & 0.36 & 0.27 & 0.24 \\
        \textbf{GRASP} &  & \underline{0.90} & \textbf{0.90} & \textbf{0.70} \\

        \bottomrule
    \end{tabular}
}
    \caption{Recall metrics for retrieved passages across multi-hop benchmarks. For Search-R1, P stands for PPO and G stands for GRPO. The best method is shown in \textbf{bold}, and the second-best method is \underline{underlined}.}
    \label{tab:ret}
\end{table}

\begin{table*}[t!]
    \centering
    \small
    \scalebox{.93}{
    \begin{tabular}{c c c c ccc ccc ccc}
        \toprule
        \multirow{2.5}{*}{\textbf{Method}}
        & \multirow{2.5}{*}{\textbf{Retriever}}
        & \multirow{2.5}{*}{\textbf{Granularity}}
        & \multirow{2.5}{*}{\textbf{Policy Learning}}
        & \multicolumn{3}{c}{\textbf{HotpotQA}}
        & \multicolumn{3}{c}{\textbf{2WikiMultihopQA}}
        & \multicolumn{3}{c}{\textbf{MuSiQue}} \\
        \cmidrule(lr){5-7}
        \cmidrule(lr){8-10}
        \cmidrule(lr){11-13}
        &  &  & 
        & EM & F1 & JD
        & EM & F1 & JD
        & EM & F1 & JD \\
        \midrule


        \multirow{3}{*}{Single-step} & Lexical & \multirow{3}{*}{Paragraph} & \multirow{3}{*}{Prompting} & 0.28 & 0.37 & 0.43 &  0.21& 0.26 &0.27  & 0.05 & 0.11 & 0.12 \\
         & Semantic &  &  & 0.30 & 0.42 & 0.50 & 0.26 & 0.31 & 0.34 & 0.10 & 0.15 & 0.20 \\
         & Hybrid &  &  & 0.37 & 0.48 & 0.56 & 0.27 & 0.32 & 0.35 & 0.13 & 0.20 & 0.23 \\
        \midrule


        IRCoT & Lexical & Paragraph & Prompting & 0.24 & 0.36 & \underline{0.63} & 0.27 & 0.40  & \underline{0.58} & 0.08 & 0.15 & 0.25 \\
        \multirow{2}{*}{Search-R1} & \multirow{2}{*}{Semantic} & \multirow{2}{*}{Paragraph} & RL (PPO) & 0.37 & 0.47 & 0.50 & 0.42 & 0.48 & 0.49 & 0.18 & 0.26 & 0.26 \\
        &  &  & RL (GRPO) & \underline{0.45} & \underline{0.56} & 0.58 & \underline{0.45} & \underline{0.53} & 0.54 & \underline{0.22} & \underline{0.30} & \underline{0.29} \\
        Base & \multirow{2}{*}{Hybrid} & \multirow{2}{*}{Sentence} & Prompting & 0.17 & 0.24 & 0.27 & 0.15 & 0.20 & 0.21 & 0.04 & 0.10 & 0.10 \\
        \textbf{GRASP} &  &  & RL (GRPO) & \textbf{0.53} & \textbf{0.66} & \textbf{0.71} & \textbf{0.52} & \textbf{0.60} & \textbf{0.63} & \textbf{0.23} & \textbf{0.33} & \textbf{0.34} \\

        \bottomrule
    \end{tabular}}
    \caption{QA Metrics across multi-hop benchmarks. Best method is \textbf{bolded} and second best is \underline{underlined}.}
    \label{tab:gen}
\end{table*}

\paragraph{Retrieval} 
Table~\ref{tab:ret} shows retrieval recall across three multi-hop QA benchmarks. Single-step results show that complementary retrieval signals are generally beneficial. Lexical retrieval alone performs substantially worse than semantic or hybrid retrieval, suggesting that lexical matching signals are often insufficient to recover all evidence required for multi-hop reasoning.
Multi-step results further show the benefit of agentic retrieval for multi-hop reasoning. With more steps, the agent can generate intermediate queries that retrieve additional evidence that may be missed by single-step retrieval. Our method achieves the strongest overall performance across the three benchmarks, suggesting that our agent effectively interleaves complementary search tools and adjusts evidence granularity as needed, enabling it to recover gold passages more reliably.

IRCoT also performs well, ranking second behind our method despite using only a lexical retriever. This result suggests that CoT-guided retrieval with a strong proprietary model, \texttt{gpt5-mini}, can generate effective intermediate queries. However, given that our agent is based on a much smaller 3B open-weight model, its stronger average recall highlights the effectiveness of learning an adaptive retrieval policy over relying on another model for CoT generation. 
Search-R1, which also learns a retrieval policy using a 3B open-weight model, performs worse than our method. Its limited performance is mainly due to incremental errors, where hallucinated intermediate queries lead subsequent queries in the wrong direction. This issue is more apparent in Base, which uses the same prompt as our method but without policy learning, resulting in significant performance decrease compared to single-step. This result highlights the importance of learning a reliable retrieval policy.

\paragraph{QA}

Table~\ref{tab:gen} shows generation quality across the three multi-hop QA benchmarks. Among single-step methods, hybrid retrieval consistently outperforms lexical and semantic retrieval, indicating that complementary retrieval signals provide more useful evidence for answer generation. However, single-step hybrid remains below the best multi-step methods, suggesting it is insufficient for multi-hop QA, where the model must recover and combine evidence across multiple supporting passages.

In the multi-step setting, GRASP performs best overall across datasets and metrics. IRCoT gets good JD scores in most cases, but its EM and F1 scores are lower, likely because it does not involve policy learning. This difference is especially clear when comparing IRCoT with Search-R1: Search-R1 generally scores higher on EM and F1, whereas IRCoT often scores higher JD. Overall, these results show that our method of learning an adaptive retrieval policy over complementary tools enables more reliable evidence acquisition and leads to stronger downstream QA performance.

\subsection{Qualitative Analysis}
\label{sec:qual}
We conduct a qualitative analysis to better understand the retrieval behaviors induced by our learned policy. Since the policy is optimized solely via RL on trajectory-level rewards without expert demonstrations, we can qualitatively analyze learned tool use trajectories, which will enable us to see what retrieval strategies emerge from the RL process.



We model the agent trajectories as a first-order Markov chain over observable actions across benchmarks as shown in Appendix Figure~\ref{fig:transition-graph}. We count all consecutive action pair 2-grams and investigate the most frequent transitions, normalized by how often each action occurs. 
%
The agent often begins with semantic search $\tau_s$ to locate a relevant topic or entity, then chooses a paragraph to read using $\tau_r$ to verify local context and extract bridge entities. These bridge entities are subsequently used in keyword search $\tau_k$, indicating a shift from broad semantic exploration to more targeted lexical retrieval. Interestingly, these patterns resemble human information foraging: readers often first skim to grasp the overall gist and form a coarse sense of relevance, then inspect promising passages in greater detail, and finally scan for specific cues~\cite{grellet1981developing}. 


After this lexical refinement step, the trajectories reveal two possible continuations.
When $\tau_k$ does not retrieve sufficient evidence, often because the keyword query is too narrow or operates at too fine a granularity, the agent returns to $\tau_s$ with a broader semantic query.
Conversely, when $\tau_k$ retrieves a promising candidate, the agent invokes $\tau_r$ to verify the relevant fact in context rather than relying solely on the search-result snippet.

We revisit the example in Figure~\ref{fig:motivation} to illustrate this behavior.
The agent first uses $\tau_s$ to search broadly for the relation among Eminem, vocals, and \textit{Unapologetic}.
Although the initial semantic result contains the distractor \textit{Encore}, the agent does not immediately advance to the next hop.
Instead, it uses $\tau_k$ with the lexically constrained query ``Eminem vocals Unapologetic'' and retrieves the bridge evidence: \textit{Unapologetic} is associated with Rihanna, and \textit{The Monster} is an Eminem song featuring guest vocals from Rihanna, taken from \textit{The Marshall Mathers LP 2}. The agent then uses $\tau_r$ to read the parent paragraph of the \textit{Unapologetic} snippet and verify that the album indeed belongs to Rihanna, establishing the first hop, \textit{Unapologetic} $\rightarrow$ Rihanna.
In a similar manner, the agent proceeds to the second hop by alternating between semantic exploration and lexical refinement around the grounded bridge entity, ultimately retrieving explicit evidence that links Rihanna to \textit{The Monster} and \textit{The Marshall Mathers LP 2}.
This trajectory shows that the learned policy uses $\tau_s$ for broad relational exploration, $\tau_k$ for lexical disambiguation around salient entities, and $\tau_r$ for contextual verification before advancing to the next hop.
As a result, it avoids the premature commitment errors made by the single-step and Search-R1 baselines (See Appendix Table~\ref{tab:naive},\ref{tab:search-r1}, and \ref{tab:grasp_eminem} for full trajectories).

\subsection{Ablation Study}
\begin{table}[t]
\centering
\small
\setlength{\tabcolsep}{8pt}
\renewcommand{\arraystretch}{1.15}
\scalebox{.95}{
\begin{tabular}{lcccc}
\toprule
\textbf{Variant} & \textbf{EM} & \textbf{$\Delta$EM} & \textbf{F1} & \textbf{$\Delta$F1} \\
\midrule
\textbf{GRASP}
    & \textbf{0.510} & -- 
    & \textbf{0.631} & -- \\
\quad without $\tau_k$
    & 0.498 & $-0.012$
    & 0.621 & $-0.010$ \\
\quad without $\tau_s$
    & 0.438 & $-0.072$
    & 0.567 & $-0.064$ \\
\quad without $\tau_r$
    & 0.388 & $-0.122$
    & 0.484 & $-0.147$ \\
\bottomrule
\end{tabular}}
\caption{
Ablation of search tools and context granularity, trained on a subset of the original HotpotQA training set. Without $\tau_r$ denotes fixed paragraph-level granularity.}
\label{tab:abl}
\end{table}

We ablate the available search tools and context granularity to evaluate whether an agent benefits from these design choices. Due to the computational cost of training multi-step retrieval agents on the full HotpotQA training set, we use a randomly sampled subset of $5{,}500$ questions and train each model for approximately two epochs, until validation performance converges across settings. The system prompt for each variant lists only the tools the agent can actually call (see Appendix Tables~\ref{tab:prompt_no_keyword}, ~\ref{tab:prompt_no_semantic}, and~\ref{tab:prompt_no_rc_paragraph} for  for the modified prompts). All other training and evaluation configurations are kept fixed. For the single-search ablations, we set the complementary search reward $R_S$ to zero, since these variants have access to only one retrieval signal and therefore cannot satisfy the reward criterion requiring both semantic and keyword search to retrieve gold evidence. For the context-granularity ablation, we set the grounded reading reward $R_R$ to zero since we remove $\tau_r$ (there are no read actions to score) and the search tools retrieve full paragraphs instead; the complementary search reward $R_S$ is retained so both search tools are still encouraged to retrieve gold evidence.

The ablation results reveal several findings. First, complementary search  contribute to performance: dropping either $\tau_k$ or $\tau_s$ reduces both EM and F1 relative to the agent equipped with both search tools, suggesting that the policy exploits both modalities rather than treating one as redundant. Second, the performance drops are asymmetric: removing semantic search leads to a substantially larger degradation than removing keyword search. 
This asymmetry aligns with the emergent behavior discussed in Section~\ref{sec:qual}, where $\tau_s$ supports broad, concept-level exploration. Without $\tau_s$, the agent has difficulty finding relevant evidence that is semantically related to the question but does not share its exact wording.
Third, context granularity impacts the quality of the agent’s intermediate queries. When the agent retrieves full paragraphs directly, it tends to generate follow-up queries that remain close to the original question, with only small edits such as appending a newly retrieved entity. Because paragraph-level results expose many entities and relations at once, the agent has a weaker signal for which specific clue should drive the next search. The intermediate query therefore becomes less targeted: instead of isolating the next missing hop, it often mixes question entities, answer-type terms, and plausible but unverified candidates. 

\section{Conclusion and Future Work}
\label{sec:conclusion}

In this paper, we studied how to train agentic RAG to make adaptive retrieval decisions and coordinate multiple tools. We introduced \textbf{GRASP}, an RL framework that trains an agent to coordinate semantic search, keyword search, and paragraph reading during multi-step reasoning. Through rewards for answer accuracy, grounded reading, complementary search, and turn efficiency, GRASP learns to retrieve evidence while controlling tool use and context expansion. Experiments on HotpotQA, 2WikiMultiHopQA, and MuSiQue show that GRASP improves retrieval recall and QA performance over single-step retrieval, prompting-based agentic RAG, and RL-based agentic RAG. Our qualitative and ablation analyses further show that the learned policy develops skimming and scanning behavior, using semantic search for broad exploration, paragraph reading for local verification, and keyword search for entity-specific evidence.

Future work could explore three directions. First, agentic RAG systems may benefit from richer tool spaces that support both evidence acquisition and reasoning, requiring studies of which tools help at different reasoning stages. Second, reward designs should reduce reliance on gold supporting-paragraph annotations by using weaker or model-based signals for evidence grounding and tool complementarity. Third, future systems should investigate dynamic retrieval granularity, allowing agents to decide how much context to expose, from short spans to paragraphs or larger document regions, based on their reasoning state.

\clearpage
\section*{Limitations}
There are several practical limitations to our work. First, our reward design depends on gold supporting-fact annotations. The grounded reading reward $R_R$ and the complementary search reward $R_S$ both require ground-truth supporting facts for scoring, which limits the framework to datasets that provide such annotations. Future work should explore weakly supervised or self-supervised alternatives, such as surrogate evidence signals derived from answer-conditioned attribution or self-consistency across trajectories, so that multi-tool reinforcement learning can be applied to corpora without per-question gold-evidence labels.
Second, the learned policy remains vulnerable to conservative termination and incomplete-retrieval failures. Although tool-call usage increases during training and stabilizes at approximately eight calls per trajectory in our final run, the policy still occasionally commits to an answer before the required evidence has been surfaced. Future work should therefore investigate reward shaping that penalizes premature commitment when the gold-evidence set has not yet been retrieved, as well as auxiliary self-assessment signals that allow the model to abstain rather than guess. Third, our experiments are limited to a single 3B-parameter backbone, \texttt{Qwen2.5-3B-Instruct}. It remains unclear how the observed emergence dynamics, including the rise of keyword search usage and the stability of the four-step retrieval pattern, depend on model scale. Larger backbones may learn the pattern faster, but they may also collapse into overly aggressive strategies that skip evidence reading. A systematic comparison across model sizes, such as 1B, 3B, 7B, and 14B, under identical reward and tool configurations would help characterize the relationship between scale and tool-use emergence.

\section*{Ethical Considerations}
There are several potential ethical considerations related to privacy, fairness, factuality, accountability, security, and appropriate human oversight. As agentic RAG systems let agents reason about their own actions, iteratively retrieve information, and potentially invoke external tools, they may introduce risks beyond those of conventional retrieval-augmented generation systems. For example, such systems may expose sensitive information from retrieved documents, amplify biases present in the underlying corpus, generate misleading outputs when retrieval results are incomplete or unreliable, or take actions that exceed the user’s intent.

To mitigate these risks, agentic RAG systems should incorporate strict access controls, data minimization, provenance tracking, citation mechanisms, and audit logs for retrieval and tool-use decisions. They should also be evaluated not only for task performance, but also for factual grounding, robustness to malicious or irrelevant retrieved content, fairness across user groups and topics, and privacy leakage. In high-impact domains, consequential actions or outputs should remain subject to human review and approval.

\section*{AI Policy}
We used AI tools including  Claude Code, Antigravity, Gemini, ChatGPT, GitHub Copilot, NotebookLM, and SciSummary to assist research design, hypothesis formulation, method implementation, literature synthesis, and assist writing. These tools supported tasks such as exploring related work, summarizing research papers, organizing notes, and refining implementation ideas. All AI-generated outputs were reviewed and edited before inclusion.

\bibliography{custom}

\appendix

\section*{Appendix}
\label{sec:appendix}

\section{Related Work}
\label{sec:related_work}
In this section, we discuss prior work related to the three core challenges addressed in this paper: retrieval-signal selection, context granularity, and learning retrieval policy using RL. To the best of our knowledge, this is the first work to study these three aspects jointly in training based   agentic RAG.

\paragraph{Lexical/Dense Retrieval}

Recent agentic RAG methods interleave reasoning with retrieval, allowing the model to generate search queries, inspect retrieved evidence, and iteratively refine its answer. In these systems, the retrieval interface is often limited to either one retrieval, such as sparse lexical retrieval~\cite{jiang2023active,trivedi2023interleaving,jeong2024adaptive}, dense semantic retrieval~\cite{li2025r3,jin2025search,leng2025decex}. As a result, the learned behavior typically focuses on when to retrieve, how to formulate queries, or how to reason over retrieved evidence, while the choice of retrieval signal itself remains largely predefined. Although hybrid retrieval has shown that lexical and semantic signals can provide complementary benefits in static RAG settings~\cite{glass2022re2g,ram2023context}, existing agentic RAG systems largely lack a mechanism for dynamically selecting and coordinating these retrieval signals during multi-step reasoning. This limitation prevents the agent from adapting its retrieval strategy to different reasoning states, such as using semantic search for broad exploration and lexical search for exact entity or fact lookup.

\paragraph{Context Granularity}

RAG performance is also strongly influenced by the granularity at which information is indexed and retrieved. Since documents are typically segmented into chunks before retrieval, chunking decisions affect both the relevance of retrieved evidence and the amount of contextual noise passed to the generator. While chunking has been extensively studied in single-step RAG pipelines~\cite{chen2024dense,wang2025document,zhao2025moc,qu2025semantic}, its role in agentic RAG remains less fully explored. Existing work often performs semantic retrieval over fixed-size or semantically segmented chunks, implicitly assuming a single level of retrieval granularity. Recent agentic approaches have begun to expose granularity as part of the retrieval process: for example, \citet{du2026rag} defines tools that allow an agent to inspect information at multiple levels, from keywords to sentences to chunks, while preserving sentence boundaries within chunked passages. However, such methods still rely primarily on dense retrieval and do not fully address how agents should choose between different retrieval signals or coordinate evidence across granularities. 


\paragraph{Learning Retrieval Policies}

Recent work has explored RL as a way to train LLM agents for iterative retrieval in multi-hop reasoning. \citet{jin2025search} shows that RL can teach an LLM to interleave reasoning with multi-turn search, but its formulation provides access to only a single retrieval tool and optimizes retrieval behavior mainly through an outcome-level answer reward. More recently, \citet{luo2025marag} extends RL-based agentic retrieval to a multi-tool setting, incorporating semantic search, keyword search, filtering, and aggregation tools, as well as rewards for answer correctness, document coverage, and tool exploration. However, its training pipeline relies on supervised fine-tuning (SFT) before RL, and its retrieved evidence remains at the chunk level, leaving unresolved the problem of how to manage evidence granularity and context-window usage efficiently.

\section{Reward Implementation Details}
\label{app:reward-impl}
This section makes concrete the abstract evidence sets used in Section~\ref{sec:reward}. HotpotQA (distractor) annotates each question with a list of \emph{supporting facts}: pairs $(t,s)$, where $t$ is the title of a gold Wikipedia article and $s$ is a gold supporting sentence in that article. In our implementation, we use the normalized Wikipedia \emph{title} as the identifier for the parent document/paragraph associated with each retrieved or read sentence. Thus, the title-level sets below instantiate the abstract document sets $\mathcal{D}_r$, $\mathcal{D}_s$, $\mathcal{D}_k$, and $\mathcal{D}_g$ used in Section~\ref{sec:reward}.

\paragraph{String normalization.}
A common normalizer $\mathrm{norm}(\cdot)$ is applied to all strings entering the reward: titles for $R_R$ and $R_S$, and answer strings for $R_A$ before token-level F1 is computed. The normalizer lowercases text, removes the articles $\{a, an, the\}$, strips punctuation, and collapses whitespace. Using a single normalizer across components ensures that surface-form variation such as case, articles, and punctuation does not create spurious mismatches between the policy outputs and the gold annotations.

\paragraph{Gold evidence set.}
Given the supporting facts, we form
\[
\mathcal{D}_g
=
\big\{
\mathrm{norm}(t) : (t,\cdot)\in\text{supporting facts}
\big\}.
\]
Duplicate titles are removed during preprocessing, so multiple supporting facts from the same Wikipedia article contribute a single evidence item.

\paragraph{Read evidence set.}
Each $\tau_r$ reads parent paragraph of a previously retrieved sentence. The backing paragraph carries a Wikipedia title $t$. We collect these titles into
\[
\mathcal{D}_r
=
\big\{
\mathrm{norm}(t_k) : k \in K_{\mathrm{read}}
\big\},
\]
where $K_{\mathrm{read}}$ indexes the paragraphs read during the trajectory and $t_k$ denotes the title of chunk $k$. We compute
\[
R_R
=
\mathrm{F1}
\big(
\mathcal{D}_r,
\mathcal{D}_g
\big).
\]
Precision is the fraction of read titles that are gold, penalizing distractor reads, and recall is the fraction of gold titles that are read, penalizing missed evidence.


\paragraph{Surfaced evidence sets.}
Both retrievers return ranked sentence-level evidence together with metadata for each sentence's parent paragraph, including a chunk identifier and title. We track surfaced evidence separately for the semantic and keyword tools. Let \(\mathcal{E}_s\) denote the set of all sentence-level result entries returned by semantic search \(\tau_s\) during the trajectory, and let \(\mathcal{E}_k\) denote the set of all sentence-level result entries returned by keyword search \(\tau_k\) during the trajectory. We then extract the normalized parent title attached to each returned sentence and form
\[
D_s = \{\mathrm{norm}(t_i) : i \in \mathcal{E}_s\},
\]
\[
D_k = \{\mathrm{norm}(t_i) : i \in \mathcal{E}_k\}.
\]
These sets instantiate the abstract evidence sets \(D_s\) and \(D_k\) from Section~3.2. A title is considered surfaced by a tool once it appears in any result list returned by that tool, regardless of rank or whether the model subsequently reads it. Reading behavior is rewarded separately by \(R_R\).

\paragraph{Complementary search indicator.}
The complementary search reward is implemented as
\[
R_S\;=\;\mathbf{1}\!\left[\,\mathcal{D}_s\cap\mathcal{D}_g\neq\varnothing \;\wedge\; \mathcal{D}_k\cap\mathcal{D}_g\neq\varnothing\,\right].
\]
Thus, $R_S=1$ exactly when semantic search surfaces at least one gold title and keyword search also surfaces at least one gold title at some point in the trajectory. The same gold title may be credited to both tools; the reward does not assign winner-takes-all attribution. 

\paragraph{Answer extraction and multi-reference $R_A$.}
When a trajectory contains multiple \texttt{<answer>...</answer>} spans, $\hat{y}$ is taken from the last such span, reflecting the agent's final commitment. When the dataset provides multiple acceptable answer strings $\{y^{(1)},\dots,y^{(m)}\}$, we set
\[
R_A
=
\max_j \mathrm{F1}(\hat{y},y^{(j)}).
\]

\paragraph{Turn count for $R_E$.}
Let $T$ denote the configured maximum turn budget and let $T_{\mathrm{cur}}$ denote the number of agent steps used by the trajectory, including the terminal step that emits \texttt{<answer>...</answer>}. We implement
\[
R_E
=
\max\!\left(0,\frac{T-T_{\mathrm{cur}}}{T}\right)
\cdot
\mathbf{1}[R_A>0.5].
\]
The non-negativity clamp guards against any rollout that exceeds the configured cap. With $T=10$, a correct one-step trajectory receives $R_E=0.9$, while a correct trajectory using all $T$ turns receives $R_E=0$.

\paragraph{Spam guard.}
As an implementation safeguard for answer extraction, trajectories whose response contains more than ten opening \texttt{<answer>} tags or more than ten closing \texttt{</answer>} tags have $R_A$ capped at $0.25$. This prevents a degenerate strategy in which the policy emits many candidate answers in a single trajectory to brute-force token-level F1. This safeguard only affects the answer-accuracy component and does not change the definitions of $R_R$, $R_S$, or $R_E$.

\section{Training Procedure Details}
\label{app:training}
This section expands on the training-procedure summary in Section~\ref{sec:implementation} and reports the exact configuration of our main GRPO run.

\begin{table}[h]
\centering
\small
\setlength{\tabcolsep}{6pt}
\renewcommand{\arraystretch}{1.1}
\begin{tabular}{p{3cm}p{3.5cm}}
\toprule
\textbf{Setting} & \textbf{Value} \\
\midrule
\multicolumn{2}{l}{\emph{Data}} \\
Dataset                       & HotpotQA (distractor) \\
Training-split size           & $90{,}447$ examples \\
Optimization steps trained    & $270$ \\
Prompts seen ($\approx$ epoch share) & $\sim\!17{,}280$ ($\sim\!19\%$) \\
Validation set                & $256$ prompts \\
\midrule
\multicolumn{2}{l}{\emph{Backbone}} \\
Policy / reference model      & \texttt{Qwen/Qwen2.5-3B-Instruct} \\
Adapter                       & none (full-parameter) \\
Parallelism                   & FSDP~\citep{zhao2023pytorch} \\
\midrule
\multicolumn{2}{l}{\emph{GRPO and rollout}} \\
Rollouts per prompt ($G$)     & $4$ \\
Prompts per step              & $64$ \\
Trajectories per step         & $256$ \\
Generation temperature        & $1.0$ \\
Max turns per trajectory ($T$)        & $10$ \\
Max response length           & $6{,}144$ tokens \\
Tool-call timeout / retries   & $180$\,s\,/\,$1$ \\
\midrule
\multicolumn{2}{l}{\emph{Optimization}} \\
Optimizer                     & AdamW (constant LR) \\
Learning rate                 & $1\mathrm{e}{-6}$ \\
GRPO clip range $\epsilon_c$  & $0.2$ \\
KL loss coefficient           & $1\mathrm{e}{-4}$ \\
Entropy coefficient           & $0.0021$ \\
\midrule
\multicolumn{2}{l}{\emph{Infrastructure}} \\
GPUs                          & $2\times$ NVIDIA A100 80\,GB \\
Framework                     & Verl-Tool~\citep{jiang2025verltool} \\
Rollout backend               & vLLM \\
Retrieval servers             & separate GPU node (FastAPI) \\
\bottomrule
\end{tabular}
\caption{Training configuration for our main GRPO model. The training step count ($270$) corresponds to the checkpoint reported in our results; we do not complete a full epoch over the split.}
\label{tab:train-config}
\end{table}

\begin{figure}[h]
\centering
\includegraphics[width=\linewidth]{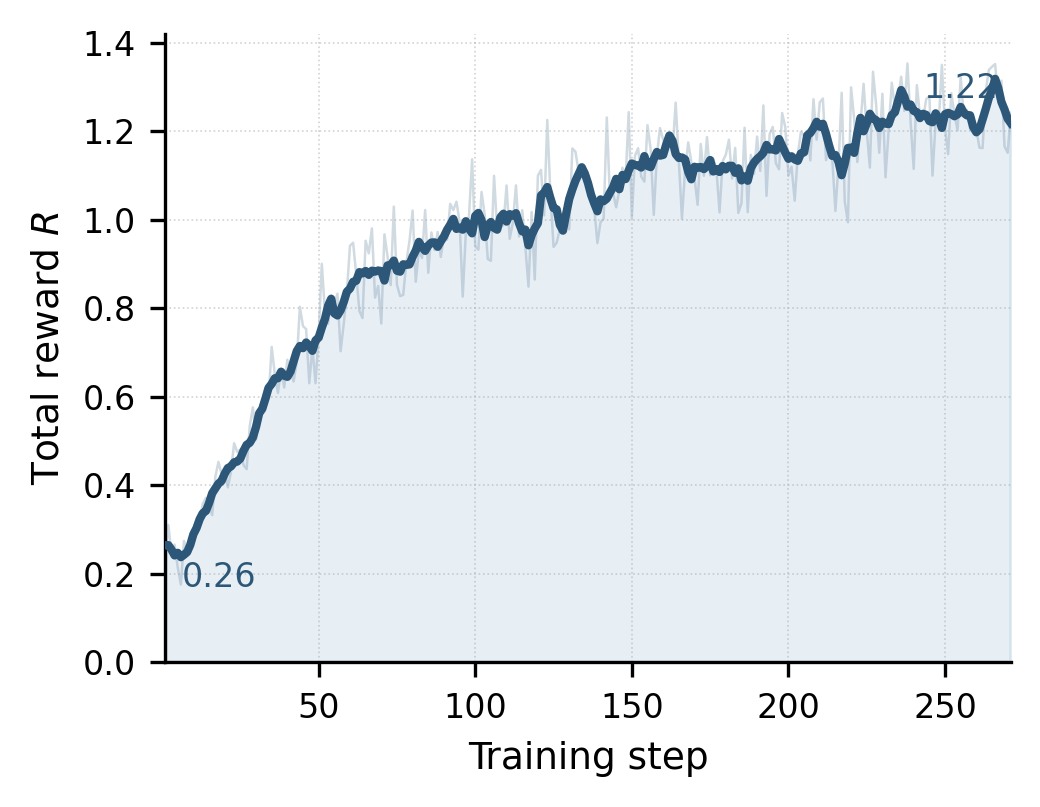}
\caption{Total trajectory-level reward $R$ over GRPO training. The bold line is a centered moving average; the faint trace behind is the raw per-step value.}
\label{fig:reward-progression}
\end{figure}

\paragraph{Training data and run length.} We train on the HotpotQA distractor training split, which contains $90{,}447$ examples. A full pass over the split at batch size $64$ corresponds to $1{,}413$ optimization steps. In practice, our main model is trained for $270$ GRPO optimization steps ($\sim\!17{,}280$ prompts, $\sim\!19\%$ of an epoch), at which point validation reward had plateaued (Figure~\ref{fig:reward-progression}); we report this checkpoint as our main model throughout the paper.

\paragraph{Loss masking.} Each rollout token is tagged at generation time as either policy-generated (model reasoning, \texttt{<semantic\_search>}, \texttt{<keyword\_search>}, \texttt{<read\_chunk>}, and \texttt{<answer>} spans) or retrieval-observation (the result list returned by a tool, inserted between turns by the rollout backend). Following \citet{jin2025search}, only the policy-generated tokens contribute to the GRPO surrogate loss: the per-token loss is multiplied by a $\{0,1\}$-valued mask aligned with these tags before being summed and normalized. This is the standard practice for tool-using RL agents and prevents the policy from receiving spurious gradients that would push it to memorize, paraphrase, or imitate the retriever's outputs.

\paragraph{Entropy schedule.} The entropy-bonus coefficient $0.0021$ in Table~\ref{tab:train-config} was selected based on training-stability ablations in which lower coefficients ($\leq 0.001$) led to premature entropy collapse during the first $\sim\!60$ optimization steps before the retrieval reward had stabilized.

\paragraph{Backbone and infrastructure.} The policy is initialized from \texttt{Qwen/Qwen2.5-3B-Instruct} and the reference policy is frozen at the same checkpoint. The model is fine-tuned with full-parameter updates using Fully Sharded Data Parallel (FSDP)~\citep{zhao2023pytorch} (no LoRA adapters). Training runs on $2\times$ NVIDIA A100 80\,GB GPUs using the Verl-Tool framework~\citep{jiang2025verltool} (a fork of verl) with vLLM~\citep{kwon2023efficient} as the rollout backend. Retrieval tools run as separate FastAPI servers on a third GPU node and are accessed by the rollout workers over HTTP; this isolation prevents the BM25 and dense-retriever working sets from competing with training memory.

\begin{figure}[ht]
\centering
\begin{tikzpicture}[
    node/.style={
        circle,
        draw=black,
        thick,
        fill=cyan!18,
        minimum size=0.5cm,
        font=\bfseries\small,
        align=center
    },
    edge/.style={
        -{Stealth[length=4mm,width=3mm]},
        draw=gray,
        opacity=0.65,
        line cap=round
    },
    prob/.style={
        text=black,
        font=\large
    }
]

\node[node] (semantic) at (0,3) {$\tau_s$};
\node[node] (read)     at (0,0) {$\tau_r$};
\node[node] (keyword)  at (6,3) {$\tau_k$};
\node[node] (answer)   at (6,0) {$\tau_a$};


\draw[edge, line width=1.5pt]
    (semantic) edge[loop above, looseness=8] node[prob, below] {0.11} (semantic);
\draw[edge, line width=2.6pt]
    (semantic) to[bend left=12] node[prob, above, pos=0.5] {0.32} (keyword);
\draw[edge, line width=3.6pt]
    (semantic) to[bend right=12] node[prob, left, rotate=90, pos=0.48] {0.53} (read);
\draw[edge, line width=0.8pt, opacity=0.25]
    (semantic) to[bend right=8] node[prob, rotate=-30, pos=0.2] {0.04} (answer);


\draw[edge, line width=3.2pt]
    (keyword) to[bend left=12] node[prob, above, pos=0.5] {0.48} (semantic);
\draw[edge, line width=3.0pt]
    (keyword) to[bend left=10] node[prob, rotate=25, pos=0.53] {0.44} (read);
\draw[edge, line width=1.1pt, opacity=0.35]
    (keyword) to[bend left=8] node[prob, right, rotate=90, pos=0.65] {0.08} (answer);


\draw[edge, line width=2.4pt]
    (read) to[bend right=12] node[prob, right, rotate=90, pos=0.52] {0.29} (semantic);
\draw[edge, line width=2.7pt]
    (read) to[bend right=-10] node[prob, rotate=35, pos=0.54] {0.35} (keyword);
\draw[edge, line width=0.7pt, opacity=0.20]
    (read) edge[loop below, looseness=8] node[prob, above] {0.01} (read);
\draw[edge, line width=2.7pt]
    (read) to[bend right=10] node[prob, above, pos=0.5] {0.35} (answer);

\end{tikzpicture}
\caption{Observed first-order Markov transition graph over agent actions.}
\label{fig:transition-graph}
\end{figure}
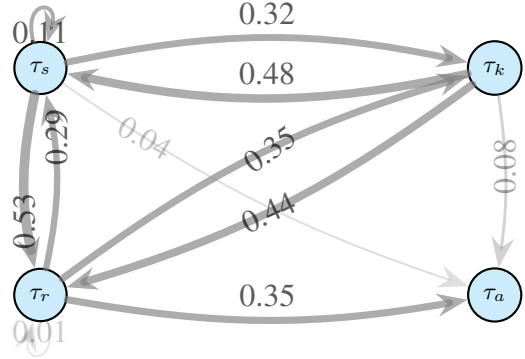

\section{Dataset Sources}
\label{app:datasets}

We obtain the three multi-hop QA datasets from the Hugging Face Hub via the \texttt{datasets} library. Table~\ref{tab:dataset-sources} lists the identifiers, the validation-split sizes used as our held-out test sets, and the size of the retrieval corpus we build for each dataset by extracting the unique supporting and distractor paragraphs across every validation question. The $500$-question evaluation subsets reported in the main paper are drawn from each validation split with \texttt{random.seed(42)}.

\paragraph{Sentence-level indexing.} Because our retriever operates at the sentence level (with parent-paragraph metadata retained for paragraph expansion), each paragraph must be split into sentences before indexing. The HotpotQA and 2WikiMultiHopQA records expose each paragraph already as a list of sentences (the \texttt{context.sentences} field, with the same schema across both datasets), so we use these sentence boundaries as-is. The MuSiQue records, in contrast, store each paragraph as a single \texttt{paragraph\_text} string, so we run NLTK's Punkt sentence tokenizer~\cite{loper2002nltk} on each paragraph to recover sentence boundaries.

\begin{table}[h]
\centering
\scriptsize
\setlength{\tabcolsep}{4pt}
\renewcommand{\arraystretch}{1.25}
\begin{tabularx}{\linewidth}{@{}l>{\raggedright\arraybackslash}Xrr@{}}
\toprule
\textbf{Dataset} & \textbf{Hugging Face identifier} & \textbf{Val.} & \textbf{Corpus} \\
\midrule
HotpotQA          & \texttt{hotpotqa/\allowbreak hotpot\_qa} (config: \texttt{distractor}) & $7{,}405$  & $66{,}635$ \\
2WikiMultiHopQA   & \texttt{framolfese/\allowbreak 2WikiMultihopQA}                        & $12{,}576$ & $56{,}687$ \\
MuSiQue           & \texttt{dgslibisey/\allowbreak MuSiQue}                                & $2{,}417$  & $21{,}100$ \\
\bottomrule
\end{tabularx}
\caption{Hugging Face dataset identifiers, validation-split sizes (used as our held-out test sets), and retrieval-corpus sizes (unique paragraphs across the full validation split) for the three multi-hop QA benchmarks used in this paper.}
\label{tab:dataset-sources}
\end{table}

\section{Agent Training Prompt}
\label{app:prompts}

This section gives the exact prompt used to train GRASP and for the ablation experiment (Table~\ref{tab:our_prompt}). The two single-tool ablations use the same template with the description block of the unavailable search tool removed (Tables~\ref{tab:prompt_no_keyword} and~\ref{tab:prompt_no_semantic}); the system prompt and the surrounding instructions are unchanged. The user message at inference time is the prompt below with the question appended in place of \texttt{\{question\}}.

\begin{table*}[t]
\centering
\small
\begin{tabular}{p{0.95\linewidth}}
\hline
\textbf{System prompt} \\
\hline
You are a helpful and harmless assistant. \\
\hline
\textbf{User prompt} \\
\hline
Answer the given question. You must conduct reasoning inside \texttt{\textless{}think\textgreater{}} and \texttt{\textless{}/think\textgreater{}} first every time you get new information. After reasoning, if you lack knowledge you can use one of these tools:

\quad \texttt{\textless{}semantic\_search\textgreater{} query \textless{}/semantic\_search\textgreater{}}.
Use when searching by concept or description --- when you are not sure of the exact name, or when you want documents related to a topic or idea. Write a descriptive natural-language phrase as the query.
Example: \texttt{\textless{}semantic\_search\textgreater{} director of the 1994 film Ed Wood \textless{}/semantic\_search\textgreater{}}.
Example: \texttt{\textless{}semantic\_search\textgreater{} founding members of Rascal Flatts \textless{}/semantic\_search\textgreater{}}.

\quad \texttt{\textless{}keyword\_search\textgreater{} query \textless{}/keyword\_search\textgreater{}}.
Use when you know a specific name, title, place, or term that should appear in the document. Write the name or phrase you want to look up. Multi-word names work fine.
Example: \texttt{\textless{}keyword\_search\textgreater{} Michael Jordan \textless{}/keyword\_search\textgreater{}}.
Example: \texttt{\textless{}keyword\_search\textgreater{} Copa del Rey \textless{}/keyword\_search\textgreater{}}.
Example: \texttt{\textless{}keyword\_search\textgreater{} Scott Derrickson \textless{}/keyword\_search\textgreater{}}.

\quad \texttt{\textless{}read\_chunk\textgreater{} chunk\_id \textless{}/read\_chunk\textgreater{}}.
Fetch the FULL text of a chunk by its ID. Search results only show a short matched snippet (``Matched: \ldots''); the full paragraph often contains the specific detail you need. Always \texttt{read\_chunk} when a search result looks relevant but the snippet does not fully answer your question.
Example: a search returns ``Chunk ID: 42 \textbar{} Title: Eiffel Tower (Similarity: 0.81)'' with matched snippet ``\ldots construction began in January 1887 \ldots''. The snippet only shows one sentence; to find who designed it, read the full paragraph: \texttt{\textless{}read\_chunk\textgreater{} 42 \textless{}/read\_chunk\textgreater{}}.

Each tool returns results between \texttt{\textless{}information\textgreater{}} and \texttt{\textless{}/information\textgreater{}}. You can call any tool as many times as you want. If you find no further external knowledge needed, provide the answer inside \texttt{\textless{}answer\textgreater{}} and \texttt{\textless{}/answer\textgreater{}}, without detailed illustrations. For example, \texttt{\textless{}answer\textgreater{} Beijing \textless{}/answer\textgreater{}}. Question: \texttt{\{question\}} \\
\hline
\end{tabular}
\caption{Training, ablation and inference prompt used for GRASP.}
\label{tab:our_prompt}
\end{table*}

\begin{table*}[t]
\centering
\small
\begin{tabular}{p{0.95\linewidth}}
\hline
\textbf{System prompt} \\
\hline
You are a helpful and harmless assistant. \\
\hline
\textbf{User prompt} \\
\hline
Answer the given question. You must conduct reasoning inside \texttt{\textless{}think\textgreater{}} and \texttt{\textless{}/think\textgreater{}} first every time you get new information. After reasoning, if you lack knowledge you can use one of these tools:

\quad \texttt{\textless{}semantic\_search\textgreater{} query \textless{}/semantic\_search\textgreater{}}.
Use when searching by concept or description --- when you are not sure of the exact name, or when you want documents related to a topic or idea. Write a descriptive natural-language phrase as the query.
Example: \texttt{\textless{}semantic\_search\textgreater{} director of the 1994 film Ed Wood \textless{}/semantic\_search\textgreater{}}.
Example: \texttt{\textless{}semantic\_search\textgreater{} founding members of Rascal Flatts \textless{}/semantic\_search\textgreater{}}.

\quad \texttt{\textless{}read\_chunk\textgreater{} chunk\_id \textless{}/read\_chunk\textgreater{}}.
Fetch the FULL text of a chunk by its ID. Search results only show a short matched snippet (``Matched: \ldots''); the full paragraph often contains the specific detail you need. Always \texttt{read\_chunk} when a search result looks relevant but the snippet does not fully answer your question.
Example: a search returns ``Chunk ID: 42 \textbar{} Title: Eiffel Tower (Similarity: 0.81)'' with matched snippet ``\ldots construction began in January 1887 \ldots''. The snippet only shows one sentence; to find who designed it, read the full paragraph: \texttt{\textless{}read\_chunk\textgreater{} 42 \textless{}/read\_chunk\textgreater{}}.

Each tool returns results between \texttt{\textless{}information\textgreater{}} and \texttt{\textless{}/information\textgreater{}}. You can call any tool as many times as you want. If you find no further external knowledge needed, provide the answer inside \texttt{\textless{}answer\textgreater{}} and \texttt{\textless{}/answer\textgreater{}}, without detailed illustrations. For example, \texttt{\textless{}answer\textgreater{} Beijing \textless{}/answer\textgreater{}}. Question: \texttt{\{question\}} \\
\hline
\end{tabular}
\caption{Prompt for the ``without $\tau_k$'' ablation. Identical to Table~\ref{tab:our_prompt} with the \texttt{keyword\_search} description block removed; the corresponding action stop token is also dropped from the rollout backend.}
\label{tab:prompt_no_keyword}
\end{table*}

\begin{table*}[t]
\centering
\small
\begin{tabular}{p{0.95\linewidth}}
\hline
\textbf{System prompt} \\
\hline
You are a helpful and harmless assistant. \\
\hline
\textbf{User prompt} \\
\hline
Answer the given question. You must conduct reasoning inside \texttt{\textless{}think\textgreater{}} and \texttt{\textless{}/think\textgreater{}} first every time you get new information. After reasoning, if you lack knowledge you can use one of these tools:

\quad \texttt{\textless{}keyword\_search\textgreater{} query \textless{}/keyword\_search\textgreater{}}.
Use when you know a specific name, title, place, or term that should appear in the document. Write the name or phrase you want to look up. Multi-word names work fine.
Example: \texttt{\textless{}keyword\_search\textgreater{} Michael Jordan \textless{}/keyword\_search\textgreater{}}.
Example: \texttt{\textless{}keyword\_search\textgreater{} Copa del Rey \textless{}/keyword\_search\textgreater{}}.
Example: \texttt{\textless{}keyword\_search\textgreater{} Scott Derrickson \textless{}/keyword\_search\textgreater{}}.

\quad \texttt{\textless{}read\_chunk\textgreater{} chunk\_id \textless{}/read\_chunk\textgreater{}}.
Fetch the FULL text of a chunk by its ID. Search results only show a short matched snippet (``Matched: \ldots''); the full paragraph often contains the specific detail you need. Always \texttt{read\_chunk} when a search result looks relevant but the snippet does not fully answer your question.
Example: a search returns ``Chunk ID: 42 \textbar{} Title: Eiffel Tower (Similarity: 0.81)'' with matched snippet ``\ldots construction began in January 1887 \ldots''. The snippet only shows one sentence; to find who designed it, read the full paragraph: \texttt{\textless{}read\_chunk\textgreater{} 42 \textless{}/read\_chunk\textgreater{}}.

Each tool returns results between \texttt{\textless{}information\textgreater{}} and \texttt{\textless{}/information\textgreater{}}. You can call any tool as many times as you want. If you find no further external knowledge needed, provide the answer inside \texttt{\textless{}answer\textgreater{}} and \texttt{\textless{}/answer\textgreater{}}, without detailed illustrations. For example, \texttt{\textless{}answer\textgreater{} Beijing \textless{}/answer\textgreater{}}. Question: \texttt{\{question\}} \\
\hline
\end{tabular}
\caption{Prompt for the ``without $\tau_s$'' ablation. Identical to Table~\ref{tab:our_prompt} with the \texttt{semantic\_search} description block removed; the corresponding action stop token is also dropped from the rollout backend.}
\label{tab:prompt_no_semantic}
\end{table*}

\begin{table*}[t]
\centering
\small
\begin{tabular}{p{0.95\linewidth}}
\hline
\textbf{System prompt} \\
\hline
You are a helpful and harmless assistant. \\
\hline
\textbf{User prompt} \\
\hline
Answer the given question. You must conduct reasoning inside \texttt{\textless{}think\textgreater{}} and \texttt{\textless{}/think\textgreater{}} first every time you get new information. After reasoning, if you lack knowledge you can use one of these tools:

\quad \texttt{\textless{}semantic\_search\textgreater{} query \textless{}/semantic\_search\textgreater{}}.
Use when searching by concept or description --- when you are not sure of the exact name, or when you want documents related to a topic or idea. Write a descriptive natural-language phrase as the query.
Example: \texttt{\textless{}semantic\_search\textgreater{} director of the 1994 film Ed Wood \textless{}/semantic\_search\textgreater{}}.
Example: \texttt{\textless{}semantic\_search\textgreater{} founding members of Rascal Flatts \textless{}/semantic\_search\textgreater{}}.

\quad \texttt{\textless{}keyword\_search\textgreater{} query \textless{}/keyword\_search\textgreater{}}.
Use when you know a specific name, title, place, or term that should appear in the document. Write the name or phrase you want to look up. Multi-word names work fine.
Example: \texttt{\textless{}keyword\_search\textgreater{} Michael Jordan \textless{}/keyword\_search\textgreater{}}.
Example: \texttt{\textless{}keyword\_search\textgreater{} Copa del Rey \textless{}/keyword\_search\textgreater{}}.
Example: \texttt{\textless{}keyword\_search\textgreater{} Scott Derrickson \textless{}/keyword\_search\textgreater{}}.

Each tool returns results between \texttt{\textless{}information\textgreater{}} and \texttt{\textless{}/information\textgreater{}}. You can call any tool as many times as you want. If you find no further external knowledge needed, provide the answer inside \texttt{\textless{}answer\textgreater{}} and \texttt{\textless{}/answer\textgreater{}}, without detailed illustrations. For example, \texttt{\textless{}answer\textgreater{} Beijing \textless{}/answer\textgreater{}}. Question: \texttt{\{question\}} \\
\hline
\end{tabular}
\caption{Prompt for the ``without $\tau_r$'' (paragraph-granularity) ablation. Identical to Table~\ref{tab:our_prompt} with the \texttt{read\_chunk} description block removed; the corresponding action stop token is also dropped from the rollout backend. The underlying retrieval index is paragraph-granular, so the search tools return the full paragraph text in each result rather than a matched-sentence snippet.}
\label{tab:prompt_no_rc_paragraph}
\end{table*}

\begin{table*}[t]
\centering
\begin{tabular}{p{0.92\linewidth}}
\hline
\textbf{System prompt} \\
\hline
You are a strict QA evaluator. For each sample, respond with ONLY ``CORRECT'' or ``INCORRECT''. \\
\hline
\textbf{Prompt} \\
\hline
You are an expert evaluator. Please evaluate if the generated answer is correct by comparing it with the gold answer.

Generated answer: \texttt{\{predicted\}}

Gold answer: \texttt{\{ground\_truth\}}

The generated answer should be considered correct if it:
1. Contains the key information from the gold answer
2. Is factually accurate and consistent with the gold answer
3. Does not contain any contradicting information \\
\hline
\end{tabular}
\caption{Prompt used for LLM-as-a-judge evaluation.}
\label{tab:judge-prompt}
\end{table*}

\onecolumn

\begin{longtable}{|p{0.95\textwidth}|}
\caption{Despite retrieving two gold documents (Doc 3 and Doc 5) for the query, the single-step model hallucinates and provides the wrong answer, ``Encore.''} \label{tab:naive}\\
\hline
\endfirsthead

\multicolumn{1}{c}%
{{\tablename\ \thetable{} -- Continued from previous page}} \\
\hline
\endhead

\hline \multicolumn{1}{r}{{Continued on next page}} \\
\endfoot

\hline
\endlastfoot

Retrieved context: \\
Doc 1 (Title: Encore (Eminem album)) Encore (stylized as ENCORE) is the fifth studio album by American rapper Eminem. It was released by Aftermath Entertainment, Shady Records, and Interscope Records. Its release was set for November 16, 2004, but was moved up to November 12 (coincidentally, exactly eight years to the day since his debut album, ``Infinite'', was released) after the album was leaked to the Internet. ``Encore'' sold 710,000 copies in its first three days, and went on to sell over 1.5 million copies in its first two weeks of release in the United States, certified quadruple-platinum that mid-December. Nine months after its release, worldwide sales of the album stood at 11 million copies. By December 2016, the album had sold over 5 million copies in the United States and more than 23 million copies worldwide. \\
Doc 2 (Title: Encore (Eminem song)) ``Encore'' (stylized as ``ENCORE'' and sometimes known as ``Curtains Down'') is a song by rappers Eminem, 50 Cent and Dr. Dre, released in 2004 as a vinyl single in the U.S. It was the title track from the Eminem album of the same name, which was also released that year. It is the third single and final track from the album.\\
Doc 3 (Title: Numb (Rihanna song)) ``Numb'' is a song by Barbadian singer Rihanna from her seventh studio album ``Unapologetic'' (2012). It features guest vocals by American rapper Eminem, making it the pair's third collaboration since the two official versions of ``Love the Way You Lie''. Following the album's release, ``Numb'' charted on multiple charts worldwide including in Canada, the United Kingdom and the United States.\\
Doc 4 (Title: List of Billboard 200 number-one albums of 2002) The highest-selling albums and EPs in the United States are ranked in the ``Billboard'' 200, published by ``Billboard'' magazine. The data are compiled by Nielsen Soundscan based on each album's weekly physical and digital sales. 25 acts achieved number one albums during this year with artist such as Nelly and Shania Twain who had their albums debut at number one on the chart. Rapper Eminem's ``The Eminem Show'' is the best selling album of 2002 selling over approximately 7.6 million copies by the end of the year. It is also the longest running album of 2002 spending six non-consecutive weeks the chart and was known for its first full week of sales debut of 1.322 million copies which Nielsen SoundScan scanned as the sixth largest sales of all time in its first week. Its debut of 1.322 million copies has still not been matched by any album today since except for Taylor Swift's album ``1989'', which opened with first week sales of 1.279 million copies. The band Creed continued its eight week long run on the chart but is credited as the longest running album 2001. Jennifer Lopez earned her second number one album on the charts with ``'', which became the highest first week sales of a remix album at the time. R\&B artist Ashanti earned her first number one album with her self-titled debut album ``Ashanti'', which opened up with first week sales of 503,000 copies in its first week alone. Puff Daddy earned his first number one album since ``No Way Out'' back in 1997. Rapper Jay-Z earned his fifth chart topper with ``'', which opened up with first week sales of 545,000 copies alone. Heavy metal band Disturbed earned its first number one album on the chart with ``Believe``, which opened up with first week sales of 284,000 copies alone. Country music singer Shania Twain's album ``Up!`` opened up with a huge first week sales of 857,000 copies in its first week alone, giving her the recognition of the highest first week sales of her career and second highest of the year, only behind Eminem's ``The Eminem Show'' and at the time the fastest selling solo female album ever. Nelly's album ``Nellyville'' opened up with his highest first week sales of his career which logged on with huge sales of 714,000 copies in its first week alone, which beat his sales of his debut album ``Country Grammar'', which opened up with first week sales of 235,000 copies. Country singer Alan Jackson album ``Drive'' gave him his first number one album on the chart and opened up with first week sales of 211,000 copies alone.\\
Doc 5 (Title: The Monster (song)) ``The Monster'' is a song by American rapper Eminem, featuring guest vocals from Barbadian singer Rihanna, taken from Eminem's album ``The Marshall Mathers LP 2'' (2013). The song was written by Eminem, Jon Bellion, and Bebe Rexha, with production handled by Frequency. ``The Monster'' marks the fourth collaboration between Eminem and Rihanna, following ``Love the Way You Lie'', its sequel ``Love the Way You Lie (Part II)'' (2010), and ``Numb'' (2012). ``The Monster'' was released on October 29, 2013, as the fourth single from the album. The song's lyrics present Rihanna coming to grips with her inner demons, while Eminem ponders the negative effects of his fame.\\
\hline
Trajectory: \\
\textless{} think\textgreater{} 
To answer this question, I need to find information about Eminem's album ``Encore'' and identify if it included a song with guest vocals from Rihanna, whose album ``Unapologetic'' we know exists. Let's check the relevant documents.

From <doc2>, we learn that ``Encore'' was the title track from Eminem's album ``Encore''. It includes a song by Eminem, 50 Cent, and Dr. Dre. 

Now, let's look at <doc3> which mentions Rihanna's song ``Numb''. This song features Eminem and was released in 2012, which aligns with the timeframe of ``Encore''.

Since ``Encore'' is the album we're looking for and it includes a song (``Numb'') with guest vocals from Rihanna, we can conclude that the album in question is ``Encore''.
\textless{}/think\textgreater{} \\
\hline
\textless{} answer\textgreater{} Encore\textless{}/answer\textgreater{} \\
\end{longtable}

\begin{longtable}{|p{0.95\textwidth}|}
\caption{Search-R1 hallucinates Kelly Clarkson as the singer, even though she is not mentioned in the retrieved documents, instead of identifying Rihanna, and incorrectly predicts ``Encore.''}\label{tab:search-r1} \\
\hline
\endfirsthead

\multicolumn{1}{c}%
{{\tablename\ \thetable{} -- Continued from previous page}} \\
\hline
\endhead

\hline \multicolumn{1}{r}{{Continued on next page}} \\
\endfoot

\hline
\endlastfoot

Trajectory: \\
To determine which Eminem album included vocals from a singer who had an album titled ``Unapologetic,'' I will follow these steps: \\
\\
1. Identify the singer who had an album titled ``Unapologetic.'' \\
2. Find out which Eminem album included vocals from this singer. \\
\\
Let's start with step 1: identifying the singer who had an album titled ``Unapologetic.'' \\
\hline
\textless{} search\textgreater{} Unapologetic album\textless{}/search\textgreater{} \\
\hline
\textless{} information\textgreater{} \\
Doc 1 (Title: Unapologetically) Unapologetically is the upcoming second studio album by American country music artist Kelsea Ballerini. It is set to be released on November 3, 2017. Ballerini announced the album's title and release date on July 25, 2017. In August, the album's track listing was first revealed to fans during a four-day check-in event over the mobile app Swarm. \\
Doc 2 (Title: No Fools, No Fun) No Fools, No Fun is the debut full-length album by Brooklyn-based alt-country trio Puss n Boots, released on July 15, 2014, through Blue Note Records. It is a collection of five original and seven cover songs originally performed by artists including Johnny Cash, Wilco and Neil Young. The album, which contains both studio and live recordings, is available on CD, vinyl and as a digital download. The album's title comes from lyrics in the Johnny Cash song, ``Bull Rider'', which is covered by the band on the album. \\
Doc 3 (Title: Reckless (Jeremy Camp album)) Reckless is contemporary Christian musician Jeremy Camp's ninth studio album, and the producers are Camp and Andy Dodd. The album was released on February 12, 2013, which all of his album has been released on BEC Recordings label, and this is his eighth album with the label. The first single from the album is entitled ``Reckless'', which was released on November 16, 2012, and has achieved chart success. Lastly, this album marks the ninth album of Camp's career that includes his first independent release. For the week of March 2, 2013, the album was the No. 31 most sold album in the nation via the ``Billboard'' 200 chart, which the album was the No. 10 most popular Rock album in the same week, and it was the most popular Christian album, as well. \\
Doc 4 (Title: Don't Lose Touch) ``Don't Lose Touch'' is a song by the Gainesville, Florida-based punk rock band Against Me!, released as the first single from their 2005 album ``Searching for a Former Clarity''. Like the second single ``From Her Lips to God's Ears (The Energizer)'', it was released exclusively on twelve-inch vinyl with a remixed version of the song as the A-side and the album version as the B-side. The A-side version was remixed by German electronic music duo Mouse on Mars. The single was limited to 3,050 copies. \\
Doc 5 (Title: Thirty Eight (album)) Thirty Eight is the fourth studio album by Detroit-based hip hop producer Apollo Brown, released digitally on April 29, 2014 by Mello Music Group. The album was later released physically, on CD and vinyl, with their own bonus tracks. Apollo Brown described ``Thirty Eight'' as a ride through Detroit of the early 1980s, that was inspired by various 1970s and 1980s films' soundtracks. One single was released from the album, ``The Answer'', as a preorder bonus. \\
\textless{}/information\textgreater{} \\
\hline
From the search results, we can see that the singer who had an album titled ``Unapologetic'' is Kelly Clarkson. Now, let's move to step 2: finding out which Eminem album included vocals from this singer. \\
\hline
\textless{} search\textgreater{} Eminem album with Kelly Clarkson vocals\textless{}/search\textgreater{} \\
\hline
\textless{} information\textgreater{} \\
Doc 1 (Title: Encore (Eminem album)) Encore (stylized as ENCORE) is the fifth studio album by American rapper Eminem. It was released by Aftermath Entertainment, Shady Records, and Interscope Records. Its release was set for November 16, 2004, but was moved up to November 12 (coincidentally, exactly eight years to the day since his debut album, ``Infinite'', was released) after the album was leaked to the Internet. ``Encore'' sold 710,000 copies in its first three days, and went on to sell over 1.5 million copies in its first two weeks of release in the United States, certified quadruple-platinum that mid-December. Nine months after its release, worldwide sales of the album stood at 11 million copies. By December 2016, the album had sold over 5 million copies in the United States and more than 23 million copies worldwide. \\
Doc 2 (Title: Encore (Eminem song)) ``Encore'' (stylized as ``ENCORE'' and sometimes known as ``Curtains Down'') is a song by rappers Eminem, 50 Cent and Dr. Dre, released in 2004 as a vinyl single in the U.S. It was the title track from the Eminem album of the same name, which was also released that year. It is the third single and final track from the album. \\
Doc 3 (Title: List of Billboard 200 number-one albums of 2002) The highest-selling albums and EPs in the United States are ranked in the ``Billboard'' 200, published by ``Billboard'' magazine. The data are compiled by Nielsen Soundscan based on each album's weekly physical and digital sales. \\
25 acts achieved number one albums during this year with artist such as Nelly and Shania Twain who had their albums debut at number one on the chart. Rapper Eminem's ``The Eminem Show'' is the best selling album of 2002 selling over approximately 7.6 million copies by the end of the year. \\
It is also the longest running album of 2002 spending six non-consecutive weeks the chart and was known for its first full week of sales debut of 1.322 million copies which Nielsen SoundScan scanned as the sixth largest sales of all time in its first week. Its debut of 1.322 million copies has still not been matched by any album today since except for Taylor Swift's album ``1989'', which opened with first week sales of 1.279 million copies. \\
The band Creed continued its eight week long run on the chart but is credited as the longest running album 2001. Jennifer Lopez earned her second number one album on the charts with `` '', which became the highest first week sales of a remix album at the time. R\&B artist Ashanti earned her first number one album with her self-titled debut album ``Ashanti'', which opened up with first week sales of 503,000 copies in its first week alone. \\
Puff Daddy earned his first number one album since ``No Way Out'' back in 1997. Rapper Jay-Z earned his fifth chart topper with `` '', which opened up with first week sales of 545,000 copies alone. Heavy metal band Disturbed earned its first number one album on the chart with ``Believe'', which opened up with first week sales of 284,000 copies alone. \\
Country music singer Shania Twain's album ``Up!'' opened up with a huge first week sales of 857,000 copies in its first week alone, giving her the recognition of the highest first week sales of her career and second highest of the year, only behind Eminem's ``The Eminem Show'' and at the time the fastest selling solo female album ever. \\
Nelly's album ``Nellyville'' opened up with his highest first week sales of his career which logged on with huge sales of 714,000 copies in its first week alone, which beat his sales of his debut album ``Country Grammar'', which opened up with first week sales of 235,000 copies. Country singer Alan Jackson album ``Drive'' gave him his first number one album on the chart and opened up with first week sales of 211,000 copies alone. \\
Doc 4 (Title: The Monster (song)) ``The Monster'' is a song by American rapper Eminem, featuring guest vocals from Barbadian singer Rihanna, taken from Eminem's album ``The Marshall Mathers LP 2'' (2013). The song was written by Eminem, Jon Bellion, and Bebe Rexha, with production handled by Frequency. ``The Monster'' marks the fourth collaboration between Eminem and Rihanna, following ``Love the Way You Lie'', its sequel ``Love the Way You Lie (Part II)'' (2010), and ``Numb'' (2012). ``The Monster'' was released on October 29, 2013, as the fourth single from the album. The song's lyrics present Rihanna coming to grips with her inner demons, while Eminem ponders the negative effects of his fame. \\
Doc 5 (Title: Imajin) Imajin is an American R\&B boy band that is known for their hit ``Shorty (You Keep Playing With My Mind)'' featuring Keith Murray. The group also made a version of this song with (rapper) Mr. Cheeks of the rap group The Lost Boyz. This single peaked at number 25 on the ``Billboard'' Hot 100, number 20 on Billboard's Hot R\&B/Hip-Hop Songs chart and number 22 in the UK Singles Chart in 1998. The band was originally put together by record producer Bert Price and inspired by past boy bands such as Jackson 5, New Edition, Hi-Five, Immature, and Mint Condition. Original credited band members included Jamal Hampton (who was later replaced by Tony Royster, Jr.), Talib Kareem, Olamide Faison, and John Fitch. Faison is the younger brother of Donald Faison of the television series ``Scrubs''. Imajin credited themselves to being a true boy band and are different because each member played an instrument. Jamal Hampton and Tony Royster played the drums, Talib Kareem the keyboard, and Olamide Faison played the guitar while John Fitch played the bass guitar. After the first album, the group split and John and Olamide made a duo called JizLams. During the hiatus, Talib continued to produce songs for various artists such as Solange Knowles. Jamal Hampton changed his name to J Star while pursuing a solo career. The group continues to produce and sing music together as ``Imajin''. \\
\textless{}/information\textgreater{} \\
\hline
From the search results, we can see that the Eminem album that included vocals from Kelly Clarkson is ``Encore.'' \\
\hline
\textless{} answer\textgreater{} Encore\textless{}/answer\textgreater{} \\
\end{longtable}

\begin{longtable}{|p{0.95\textwidth}|}
\caption{Retrieval process that first searches broadly for the connection between Eminem, vocals, and ``Unapologetic,'' then identifies Rihanna as the artist of ``Unapologetic,'' and finally confirms that Rihanna's guest vocals appear on Eminem's ``The Monster,'' which is from \textit{The Marshall Mathers LP 2}.}\label{tab:grasp_eminem} \\
\hline
\endfirsthead

\multicolumn{1}{c}%
{{\tablename\ \thetable{} -- Continued from previous page}} \\
\hline
\endhead

\hline \multicolumn{1}{r}{{Continued on next page}} \\
\endfoot

\hline
\endlastfoot

Trajectory: \\
To determine which Eminem album included vocals from a singer who had an album titled ``Unapologetic,'' I will follow these steps: \\
\\
1. Search for information connecting Eminem, vocals, and the album title ``Unapologetic.'' \\
2. Identify the singer associated with the album ``Unapologetic.'' \\
3. Find which Eminem album included vocals from that singer. \\
\\
Let's start by searching for an Eminem album that included vocals from a singer who had an album titled ``Unapologetic.'' \\
\hline
\textless{} semantic\_search\textgreater{} Eminem album included vocals from a singer who had an album titled ``Unapologetic'' \textless{}/semantic\_search\textgreater{} \\
\hline
\textless{} information\textgreater{} \\
Chunk ID: 1325 \textbar{} Title: Encore (Eminem song) (Similarity: 0.719) \\
Matched: ... It was the title track from the Eminem album of the same name, which was also released that year. ... \\
\\
Chunk ID: 42881 \textbar{} Title: Love the Way You Lie (Part II) (Similarity: 0.686) \\
Matched: ... It features guest vocals from American rapper Eminem, who wrote the song alongside Skylar Grey and the producer Alex da Kid. ... \\
\\
Chunk ID: 63870 \textbar{} Title: So Fresh, So Clean (Similarity: 0.661) \\
Matched: ... It features uncredited vocals from singer Sleepy Brown. ... \\
\\
Chunk ID: 42878 \textbar{} Title: Loud (Rihanna album) (Similarity: 0.656) \\
Matched: ... The album features several guest vocalists, including rappers Drake, Nicki Minaj and Eminem, who is featured on the sequel to ``Love the Way You Lie'', titled ``Love the Way You Lie (Part II)''. ... \\
\\
Chunk ID: 46256 \textbar{} Title: Diwa de Leon (Similarity: 0.646) \\
Matched: ... His songs included in ``Emir''. ... \\
\textless{}/information\textgreater{} \\
\hline
The first search results do not directly identify the album. They mention Eminem-related tracks and Rihanna-related material, but they do not yet establish the full connection to ``Unapologetic.'' I will now search more directly for Eminem, vocals, and ``Unapologetic.'' \\
\hline
\textless{} keyword\_search\textgreater{} Eminem vocals Unapologetic \textless{}/keyword\_search\textgreater{} \\
\hline
\textless{} information\textgreater{} \\
Chunk ID: 1327 \textbar{} Title: The Monster (song) \textbar{} 6 sentences \\
Matched: ... ``The Monster'' is a song by American rapper Eminem, featuring guest vocals from Barbadian singer Rihanna, taken from Eminem's album ``The Marshall Mathers LP 2'' (2013). ... The song was written by Eminem, Jon Bellion, and Bebe Rexha, with production handled by Frequency. ... The song's lyrics present Rihanna coming to grips with her inner demons, while Eminem ponders the negative effects of his fame. ... ``The Monster'' marks the fourth collaboration between Eminem and Rihanna, following ``Love the Way You Lie'', its sequel ``Love the Way You Lie (Part II)'' (2010), and ``Numb'' (2012). ... \\
\\
Chunk ID: 42881 \textbar{} Title: Love the Way You Lie (Part II) \textbar{} 6 sentences \\
Matched: ... It features guest vocals from American rapper Eminem, who wrote the song alongside Skylar Grey and the producer Alex da Kid. ... It is the sequel to the 2010 hit single ``Love the Way You Lie'', which appears on Eminem's seventh studio album ``Recovery''. ... \\
\\
Chunk ID: 1328 \textbar{} Title: Numb (Rihanna song) \textbar{} 4 sentences \\
Matched: ... It features guest vocals by American rapper Eminem, making it the pair's third collaboration since the two official versions of ``Love the Way You Lie''. ... ``Numb'' is a song by Barbadian singer Rihanna from her seventh studio album ``Unapologetic'' (2012). ... \\
\\
Chunk ID: 65383 \textbar{} Title: Unapologetic \textbar{} 6 sentences \\
Matched: ... Unapologetic is the seventh studio album by Barbadian singer Rihanna. ... ``Unapologetic'' is mainly a pop, dubstep and R\&B album that incorporates elements of hip hop, EDM and reggae in its production, similar to the sound of her previous albums ``Talk That Talk'' and ``Rated R'' (2009). ... \\
\\
Chunk ID: 24034 \textbar{} Title: Luis Resto (musician) \textbar{} 8 sentences \\
Matched: ... He has played the keyboard for several Eminem-produced tracks and is credited for additional production on most Eminem-produced tracks on ``Encore''. ... Luis Edgardo Resto is an American musician, producer and keyboardist who has worked closely with rapper Eminem since his third major-label album ``The Eminem Show''. ... \\
\textless{}/information\textgreater{} \\
\hline
This search provides two important pieces of evidence. First, ``Unapologetic'' is associated with Rihanna. Second, ``The Monster'' is an Eminem song featuring guest vocals from Rihanna and is taken from ``The Marshall Mathers LP 2.'' To verify the singer associated with ``Unapologetic,'' I will read the chunk about that album. \\
\hline
\textless{} read\_chunk\textgreater{} 65383 \textless{}/read\_chunk\textgreater{} \\
\hline
\textless{} information\textgreater{} \\
{[}Chunk ID: 65383 \textbar{} Title: Unapologetic{]} \\
Unapologetic is the seventh studio album by Barbadian singer Rihanna. It was released on November 19, 2012, by Def Jam Recordings and SRP Records. It was recorded between June and November 2012, during promotion of her sixth album, ``Talk That Talk'' (2011). As executive producer, Rihanna enlisted previous collaborators The-Dream, David Guetta, Chase \& Status, and StarGate to work alongside new collaborators such as Parker Ighile, Mike Will Made-It, and Labrinth. ``Unapologetic'' is mainly a pop, dubstep and R\&B album that incorporates elements of hip hop, EDM and reggae in its production, similar to the sound of her previous albums ``Talk That Talk'' and ``Rated R'' (2009). \\
\textless{}/information\textgreater{} \\
\hline
Now it is confirmed that the singer who had an album titled ``Unapologetic'' is Rihanna. Next, I need to confirm which Eminem album included vocals from Rihanna. \\
\hline
\textless{} semantic\_search\textgreater{} Rihanna Unapologetic Eminem vocals \textless{}/semantic\_search\textgreater{} \\
\hline
\textless{} information\textgreater{} \\
Chunk ID: 28099 \textbar{} Title: My Way (Usher song) (Similarity: 0.680) \\
Matched: ... It features an uncredited rap and background vocals from Jermaine Dupri. ... \\
\\
Chunk ID: 1327 \textbar{} Title: The Monster (song) (Similarity: 0.677) \\
Matched: ... The song's lyrics present Rihanna coming to grips with her inner demons, while Eminem ponders the negative effects of his fame. ... \\
\\
Chunk ID: 65386 \textbar{} Title: Right Now (Rihanna song) (Similarity: 0.669) \\
Matched: ... The lyrical content features Rihanna chanting to live life in the moment. ... \\
\\
Chunk ID: 21542 \textbar{} Title: Love the Way You Lie (Similarity: 0.667) \\
Matched: ... Eminem wrote the verses and chose Rihanna to sing the chorus, resulting in a collaboration influenced by their past experiences in difficult relationships. ... Backed by guitar, piano and violin, the track is a midtempo hip hop ballad with a pop refrain, sung by Rihanna, and describes two lovers who refuse to separate despite being in a dangerous love--hate relationship. ... \\
\\
Chunk ID: 42881 \textbar{} Title: Love the Way You Lie (Part II) (Similarity: 0.664) \\
Matched: ... It features guest vocals from American rapper Eminem, who wrote the song alongside Skylar Grey and the producer Alex da Kid. ... \\
\textless{}/information\textgreater{} \\
\hline
The result again points toward ``The Monster,'' but the snippet does not fully state the album in this step. I will use a more direct keyword search combining Eminem, ``Unapologetic,'' and vocals. \\
\hline
\textless{} keyword\_search\textgreater{} Eminem Unapologetic vocals \textless{}/keyword\_search\textgreater{} \\
\hline
\textless{} information\textgreater{} \\
Chunk ID: 1327 \textbar{} Title: The Monster (song) \textbar{} 6 sentences \\
Matched: ... ``The Monster'' is a song by American rapper Eminem, featuring guest vocals from Barbadian singer Rihanna, taken from Eminem's album ``The Marshall Mathers LP 2'' (2013). ... The song was written by Eminem, Jon Bellion, and Bebe Rexha, with production handled by Frequency. ... The song's lyrics present Rihanna coming to grips with her inner demons, while Eminem ponders the negative effects of his fame. ... ``The Monster'' marks the fourth collaboration between Eminem and Rihanna, following ``Love the Way You Lie'', its sequel ``Love the Way You Lie (Part II)'' (2010), and ``Numb'' (2012). ... \\
\\
Chunk ID: 42881 \textbar{} Title: Love the Way You Lie (Part II) \textbar{} 6 sentences \\
Matched: ... It features guest vocals from American rapper Eminem, who wrote the song alongside Skylar Grey and the producer Alex da Kid. ... It is the sequel to the 2010 hit single ``Love the Way You Lie'', which appears on Eminem's seventh studio album ``Recovery''. ... \\
\\
Chunk ID: 1328 \textbar{} Title: Numb (Rihanna song) \textbar{} 4 sentences \\
Matched: ... It features guest vocals by American rapper Eminem, making it the pair's third collaboration since the two official versions of ``Love the Way You Lie''. ... ``Numb'' is a song by Barbadian singer Rihanna from her seventh studio album ``Unapologetic'' (2012). ... \\
\\
Chunk ID: 65383 \textbar{} Title: Unapologetic \textbar{} 6 sentences \\
Matched: ... Unapologetic is the seventh studio album by Barbadian singer Rihanna. ... ``Unapologetic'' is mainly a pop, dubstep and R\&B album that incorporates elements of hip hop, EDM and reggae in its production, similar to the sound of her previous albums ``Talk That Talk'' and ``Rated R'' (2009). ... \\
\\
Chunk ID: 24034 \textbar{} Title: Luis Resto (musician) \textbar{} 8 sentences \\
Matched: ... He has played the keyboard for several Eminem-produced tracks and is credited for additional production on most Eminem-produced tracks on ``Encore''. ... Luis Edgardo Resto is an American musician, producer and keyboardist who has worked closely with rapper Eminem since his third major-label album ``The Eminem Show''. ... \\
\textless{}/information\textgreater{} \\
\hline
The evidence shows that Rihanna is the singer who had an album titled ``Unapologetic.'' It also shows that Eminem's song ``The Monster'' features guest vocals from Rihanna and was taken from Eminem's album ``The Marshall Mathers LP 2.'' Therefore, the Eminem album is ``The Marshall Mathers LP 2.'' \\
\hline
\textless{} answer\textgreater{} The Marshall Mathers LP 2 \textless{}/answer\textgreater{} \\
\end{longtable}

\begin{longtable}{|p{0.95\textwidth}|}
\caption{Another trajectory of GRASP.}\label{tab:grasp_rus} \\
\hline
\endfirsthead

\multicolumn{1}{c}%
{{\tablename\ \thetable{} -- Continued from previous page}} \\
\hline
\endhead

\hline \multicolumn{1}{r}{{Continued on next page}} \\
\endfoot

\hline
\endlastfoot

Trajectory: \\
\textless{}think\textgreater{} \\
First, I need to find information about Sergei Aleksandrovich Tokarev to find out where he was a professor. Then, I'll try to find information about the university he taught at to find out its founding year. \\
\textless{}/think\textgreater{} \\
\textless{}semantic\_search\textgreater{} query: Sergei Aleksandrovich Tokarev professor university founded \textless{}/semantic\_search\textgreater{} \\
\hline

\textless{}information\textgreater{} \\
Chunk ID: 48408 | Title: Sergei Aleksandrovich Tokarev (Similarity: 0.691) \\
Matched: ... Sergei Aleksandrovich Tokarev (Russian: [Name in Russian] , 29 December 1899 – 19 April 1985) was a Russian scholar, ethnographer, historian, researcher of religious beliefs, doctor of historical sciences, and professor at Moscow State University. ... \\
\\
Chunk ID: 17611 | Title: Liubomyr Vynar (Similarity: 0.685) \\
Matched: ... Professor. ... \\
\\
Chunk ID: 51535 | Title: Eva Schönbeck-Temesy (Similarity: 0.627) \\
Matched: ... Univ. ... \\
\\
Chunk ID: 18007 | Title: Mohd Noh Dalimin (Similarity: 0.618) \\
Matched: ... Dr. Ismail Hj. ... \\
\\
Chunk ID: 27462 | Title: Jack Haig (Similarity: 0.613) \\
Matched: ... as Monsieur Roger LeClerc ... \\
\textless{}/information\textgreater{} \\
\hline

\textless{}think\textgreater{} \\
The search results do not directly provide information about the university where Sergei Aleksandrovich Tokarev was a professor. I need to find more information about Moscow State University to find its founding year. \\
\textless{}/think\textgreater{} \\
\textless{}read\_chunk\textgreater{} 48408 \textless{}/read\_chunk\textgreater{} \\
\hline

\textless{}information\textgreater{} \\
{[}Chunk ID: 48408 | Title: Sergei Aleksandrovich Tokarev{]} \\
Sergei Aleksandrovich Tokarev (Russian: [Name in Russian] , 29 December 1899 – 19 April 1985) was a Russian scholar, ethnographer, historian, researcher of religious beliefs, doctor of historical sciences, and professor at Moscow State University. \\
\textless{}/information\textgreater{} \\
\hline

\textless{}think\textgreater{} \\
Now I have found that Sergei Aleksandrovich Tokarev was a professor at Moscow State University. Next, I need to find information about when Moscow State University was founded. \\
\textless{}/think\textgreater{} \\
\textless{}keyword\_search\textgreater{} Sergei Aleksandrovich Tokarev Moscow State University founded \textless{}/keyword\_search\textgreater{} \\
\hline

\textless{}information\textgreater{} \\
Chunk ID: 48408 | Title: Sergei Aleksandrovich Tokarev | 2 sentences \\
Matched: ... Sergei Aleksandrovich Tokarev (Russian: [Name in Russian] , 29 December 1899 – 19 April 1985) was a Russian scholar, ethnographer, historian, researcher of religious beliefs, doctor of historical sciences, and professor at Moscow State University. ... \\
\\
Chunk ID: 58241 | Title: STANKIN | 4 sentences \\
Matched: ... Moscow State Technological University “Stankin“ (MSTU Stankin) (Russian: [Russian Name] ) (previously Moscow State Technological Institute STANKIN or Moscow Machine-Instrument Institute) is one of the leading Russian technical institutes founded in 1930. ... \\
\\
Chunk ID: 48405 | Title: Sergei Panchin | 2 sentences \\
Matched: ... Sergei Aleksandrovich Panchin (Russian: [Name in Russian]; born 15 December 1993) is a Russian football goalkeeper. ... \\
\\
Chunk ID: 48407 | Title: Sergei Dmitrochenko | 2 sentences \\
Matched: ... Sergei Aleksandrovich Dmitrochenko (Russian: [Name in Russian] ; born June 21, 1993) is a Russian football midfielder. ... \\
\\
Chunk ID: 48409 | Title: Sergei Aleksandrovich Kudryavtsev | 5 sentences \\
Matched: ... Sergei Aleksandrovich Kudryavtsev (1903 – April 25, 1938) was a Ukrainian communist Soviet politician. ... \\
\textless{}/information\textgreater{} \\
\hline

\textless{}think\textgreater{} \\
The search results do not provide information about the founding year of Moscow State University. I need to find more information about Moscow State University itself. \\
\textless{}/think\textgreater{} \\
\textless{}semantic\_search\textgreater{} query: Moscow State University founder \textless{}/semantic\_search\textgreater{} \\
\hline

\textless{}information\textgreater{} \\
Chunk ID: 58232 | Title: Belgorod Technological University (Similarity: 0.645) \\
Matched: ... [Name in Russian])) is the largest russian university with specialization in the field of the construction materials. ... \\
\\
Chunk ID: 18007 | Title: Mohd Noh Dalimin (Similarity: 0.641) \\
Matched: ... Dr. Ismail Hj. ... \\
\\
Chunk ID: 2073 | Title: Alexandroff extension (Similarity: 0.638) \\
Matched: ... It is named for the Russian mathematician Pavel Alexandrov. ... \\
\\
Chunk ID: 40432 | Title: Adam Clayton Powell Jr. State Office Building (Similarity: 0.635) \\
Matched: ... The Adam Clayton Powell Jr. ... \\
\\
Chunk ID: 23118 | Title: Martin Luther King Jr. Outpatient Center (Similarity: 0.627) \\
Matched: ... The Martin Luther King Jr. ... \\
\textless{}/information\textgreater{} \\
\hline

\textless{}think\textgreater{} \\
The search results do not provide information about the founding year of Moscow State University. I need to find more information about Moscow State University itself. \\
\textless{}/think\textgreater{} \\
\textless{}semantic\_search\textgreater{} query: Moscow State University history \textless{}/semantic\_search\textgreater{} \\
\hline

\textless{}information\textgreater{} \\
Chunk ID: 58241 | Title: STANKIN (Similarity: 0.627) \\
Matched: ... Moscow State Technological University “Stankin“ (MSTU Stankin) (Russian: [Name in Russian] ) (previously Moscow State Technological Institute STANKIN or Moscow Machine-Instrument Institute) is one of the leading Russian technical institutes founded in 1930. ... \\
\\
Chunk ID: 48400 | Title: Moscow State University (Similarity: 0.615) \\
Matched: ... Lomonosov Moscow State University (MSU; Russian: [Name in Russian]. ... [Name in Russian] , often abbreviated [Name in Russian]) is a coeducational and public research university located in Moscow, Russia. ... MSU was renamed after Lomonosov in 1940 and was then known as "Lomonosov University". ... \\
\\
Chunk ID: 58236 | Title: Siberian State Technological University (Similarity: 0.609) \\
Matched: ... Siberian State Technological University is the oldest university in Krasnoyarsk, Russia. ... \\
\\
Chunk ID: 13043 | Title: Vsya Rossiya (Similarity: 0.608) \\
Matched: ... Historians use them to research the social histories of late 19th century and early 20th century Russia. ... \\
\\
Chunk ID: 51535 | Title: Eva Schönbeck-Temesy (Similarity: 0.607) \\
Matched: ... Univ. ... \\
\textless{}/information\textgreater{} \\
\hline

\textless{}think\textgreater{} \\
The search results do not provide direct information about the founding year of Moscow State University. I need to find more information about Moscow State University from the relevant chunk of text. \\
\textless{}/think\textgreater{} \\
\textless{}read\_chunk\textgreater{} 48400 \textless{}/read\_chunk\textgreater{} \\
\hline

\textless{}information\textgreater{} \\
{[}Chunk ID: 48400 | Title: Moscow State University{]} \\
Lomonosov Moscow State University (MSU; Russian: [Name in Russian] , often abbreviated [Name in Russian]) is a coeducational and public research university located in Moscow, Russia. It was founded on January 25, 1755 by Mikhail Lomonosov. MSU was renamed after Lomonosov in 1940 and was then known as "Lomonosov University". It also houses the tallest educational building in the world. It is rated among the universities with the best reputation in the world. Its current rector is Viktor Sadovnichiy. \\
\textless{}/information\textgreater{} \\
\hline

\textless{}think\textgreater{} \\
Now I have found that Moscow State University was founded on January 25, 1755. Therefore, the year when the university where Sergei Aleksandrovich Tokarev was a professor was founded is 1755. \\
\textless{}/think\textgreater{} \\
\textless{}answer\textgreater{}1755\textless{}/answer\textgreater{} \\
\end{longtable}

\end{document}